\documentclass[10pt,twocolumn,letterpaper]{article}

\usepackage[pagenumbers]{iccv} 
\usepackage{multirow}
\usepackage{makecell}
\usepackage{graphicx}
\definecolor{iccvblue}{rgb}{0.21,0.49,0.74}
\usepackage[pagebackref,breaklinks,colorlinks,allcolors=iccvblue]{hyperref}

\def\paperID{5558} 
\def\confName{ICCV}
\def\confYear{2025}

\title{Fine-Grained Evaluation of Large \\ Vision-Language
Models in Autonomous Driving}
\author{
    Yue Li \textsuperscript{1} \quad
    Meng Tian \textsuperscript{2} \quad
    Zhenyu Lin \textsuperscript{2} \quad
    Jiangtong Zhu \textsuperscript{2} \quad 
    Dechang Zhu \textsuperscript{2} \quad \\
    Haiqiang Liu \textsuperscript{2} \quad
    Zining Wang \textsuperscript{3} \quad
    Yueyi Zhang \textsuperscript{1} \quad
    Zhiwei Xiong \textsuperscript{1*} \quad
    Xinhai Zhao \textsuperscript{2*}
    \\
    \textsuperscript{1} University of Science and Technology of China \\
    \textsuperscript{2} Huawei Noah’s Ark Lab \quad
    \textsuperscript{3} University of California,
Berkeley \\
    \quad
    \tt\small \{yueli65@mail., zixiong@\}ustc.edu.cn \quad \{tianmeng25, zhaoxinhai1\}@huawei.com
}

\begin{document}
\maketitle
\def\thefootnote{$*$}
\footnote{Corresponding author. The work was done during Yue Li’s internship at Huawei Noah’s Ark Lab.}
\thispagestyle{empty}
\begin{abstract}
Existing benchmarks for Vision-Language Model (VLM) on autonomous driving (AD) primarily assess interpretability through open-form visual question answering (QA) within coarse-grained tasks, which remain insufficient to assess capabilities in complex driving scenarios. To this end, we introduce $\textbf{VLADBench}$, a challenging and fine-grained dataset featuring close-form QAs that progress from static foundational knowledge and elements to advanced reasoning for dynamic on-road situations. The elaborate $\textbf{VLADBench}$ spans 5 key domains: Traffic Knowledge Understanding, General Element Recognition, Traffic Graph Generation, Target Attribute Comprehension, and Ego Decision-Making and Planning. These domains are further broken down into 11 secondary aspects and 29 tertiary tasks for a granular evaluation. A thorough assessment of general and domain-specific (DS) VLMs on this benchmark reveals both their strengths and critical limitations in AD contexts. To further exploit the cognitive and reasoning interactions among the 5 domains for AD understanding, we start from a small-scale VLM and train the DS models on individual domain datasets (collected from 1.4M DS QAs across public sources).
The experimental results demonstrate that the proposed benchmark provides a crucial step toward a more comprehensive assessment of VLMs in AD, paving the way for the development of more cognitively sophisticated and reasoning-capable AD systems. The benchmark and DS model will be available at \url{https://github.com/Depth2World/VLADBench}.

\end{abstract}    
\section{Introduction}
\label{sec:intro}

\begin{figure}[t]
  \centering
   \includegraphics[width=\linewidth]{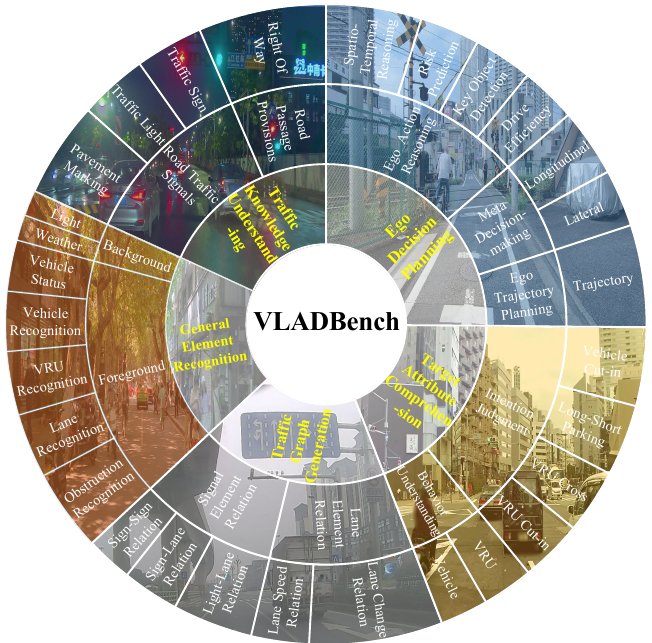}
   \vspace{-0.6cm}
   \caption{A sunburst chart of \textbf{VLADBench} categories. The proposed dataset spans 5 key domains, 11 secondary aspects and 29 tertiary tasks, including about 2,000 static scenes and 3000 dynamic scenarios, comprising 12,000 close-form questions.
   }
   \label{fig:onecol}
   \vspace{-0.6cm}
\end{figure}
Large Vision-Language Models (VLMs) are rapidly transforming numerous fields, demonstrating their potential to revolutionize how we interact with information and technology. Their ability to seamlessly integrate visual and textual data unlocks new possibilities across diverse applications, including visual content generation~\cite{liu2023llava,awadalla2023openflamingo,li2023videochat}, medical image analysis~\cite{chen2023generative,moor2023med,ye2023mplug}, robotic control~\cite{karamcheti2023language,zitkovich2023rt}, and autonomous driving (AD)~\cite{xu2024drivegpt4,ding2023hilm,ding2024holistic,tian2024drivevlm,sima2023drivelm}.


Recent VLM-based AD algorithms address the limitations of end-to-end AD approaches, including interpretability and long-tail problem, which refers to the limited generalization to new scenarios, unexpected events, and diverse traffic patterns~\cite{teng2023motion,chen2024end}. State-of-the-art models~\cite{xu2024drivegpt4,nie2025reason2drive,marcu2312lingoqa,mao2023gpt,tian2024drivevlm,sima2023drivelm} demonstrate promising results in scene perception, description, and decision-making with analysis in open form visual question answers (VQA) task. Most existing datasets concentrate on broad task categories (such as perception, prediction, planning, etc.) in AD. 


Despite recent advances, effective transfer of foundational VLMs to AD-specific models remains under-explored, in part due to the insufficient validation protocols within the context of AD. A comprehensive evaluation is necessary to guide the model transfer, focusing on strengths and weaknesses and highlighting the specific competencies that require attention, beyond merely achieving high scores on open-form VQA tasks. Current benchmarks designed for VLM-based AD face several notable limitations: 1) \textbf{Coarse-grained Categories}: The underlying datasets of the VLM-based models are often simplistic, typically categorizing tasks into perception, prediction, and planning with reasoning, which are incomplete for evaluating the nuanced cognitive and reasoning abilities required for safe and reliable AD. A holistic evaluation framework remains necessary to fully assess these critical competencies.
2) \textbf{Lack of Dynamic Elements Analysis}: 
Both static and dynamic scenes are crucial for evaluating AD systems, a robust analysis of dynamic elements is particularly important for validating the temporal reasoning capabilities, especially in understanding traffic participant intentions within the scene and executing the nuanced spatio-temporal reasoning required for safe navigation. 3) \textbf{Homogeneous Datasets}: Existing VLM-based AD datasets often suffer from a lack of diversity, which limits the ability to test models across a wide range of real-world scenarios.
The narrow results restrict the evaluation of zero-shot generalization and the performance on challenging corner cases. A more diverse dataset is required to thoroughly assess the robustness and adaptability of VLMs in real-world settings.



To overcome these limitations, we introduce a novel benchmark, \textbf{VLADBench}, specifically designed to rigorously evaluate the capabilities of VLMs in AD. \textbf{VLADBench} addresses the shortcomings of existing benchmarks by employing a hierarchical structure that reflects the complex skill set required for reliable driving, progressing from fundamental scene and traffic elements comprehension to advanced reasoning and decision-making. With 2000 static scenes and 3000 dynamic scenarios, \textbf{VLADBench} spans 5 primary domains: Traffic Knowledge Understanding (TKU), General Element Recognition (GER), Traffic Graph Generation (TGG), Target Attribute Comprehension (TAC), and Ego Decision-making and Planning (EDP). For a more detailed assessment, 11 secondary aspects and 29 tertiary
tasks are defined, resulting in a total of 12,000 questions. The dataset is built from existing publicly available datasets, meticulously curated through a manual selection across 12 sources, aimed at challenging VLM capabilities in diverse challenging driving situations. 
To further investigate the intersections among the 5 key domains, we collect and construct approximately 1.4M AD-specific QAs from public resources. We then categorize these QAs using GPT-4 and train models on individual domain-specific (DS) datasets. Finally, we validate the trained models on \textbf{VLADBench} to assess their performance across different domains.

A thorough evaluation on \textbf{VLADBench} of the prominent VLMs, encompassing both open-source (ranging from 4B to 76B), close-source and DS models, reveals the following key findings: 
\begin{itemize}
    \item Current VLMs, including the state-of-the-art large-scale
    Qwen2.5-VL-72B~\cite{wang2024qwen2}, GPT-4o, and the DS model DriveMM~\cite{huang2024drivemm}, struggle to achieve 60\% accuracy on \textbf{VLADBench}, remaining a large room for improvement.
    \item 
   Significant challenges persist especially in areas: traffic signals and graph generation, intention judgment, and meta decision-making, which are essential capabilities for achieving reliable autonomous driving. 
    \item Biased DS training data enhance the performance in certain specialized areas of autonomous driving but will lead to a loss of generalization ability in tasks that require broader and general knowledge.
    \item The DS data from the five key domains is interconnected, providing mutual benefits across domains and demonstrating a clear synergy effect.
    \item Elevating the vision encoder may be more impactful than simply scaling up the language model for AD context.
\end{itemize}

\section{Related Work}
\label{sec:related_work}

\subsection{Large Vision-Language Models}
Recent advancements in Large Language Models (LLMs) like the GPT series~\cite{achiam2023gpt} and LLaMA~\cite{touvron2023llama} have revolutionized natural language processing. This progress has spurred the development of Large Vision-Language Models, aiming to extend LLM capabilities to encompass visual understanding and reasoning. Models such as LLaVA~\cite{liu2023llava,liu2024llavanext}, MiniGPT-4~\cite{zhu2023minigpt}, InstructBLIP~\cite{instructblip}, Cambrain~\cite{tong2024cambrian}, ShareGPT4V~\cite{chen2023sharegpt4v} integrate visual information, enabling tasks like image captioning and visual question answering. These VLMs typically align visual and linguistic features using cross-attention mechanisms or MLP projections, trained on extensive image-text datasets. Early VLMs focused on static images, but recent efforts have extended their capabilities to video understanding, such as BLIP2~\cite{li2023blip2}, InternLM-XComposer2.5~\cite{zhang2023internlm}, InternVL2~\cite{chen2024internvl}, VILA~\cite{lin2024vila}, Qwen2-VL~\cite{wang2024qwen2}, etc., incorporating temporal dynamics into the language feature space for sequence comprehension. 
VLMs have demonstrated promising capabilities across diverse domains, including content creation, medical image analysis, robotics and autonomous driving.

\subsection{VLM-based Autonomous Driving}

End-to-end AD~\cite{Hu_2023_CVPR,jiang2023vad} represents a shift from traditional modular pipelines to a singular framework, which learns relevant features directly from raw sensor data and discovers effective representations with all modules training together. While models trained on specific datasets encounter the reliance on ego status ~\cite{zhai2023rethinking,li2024ego} and 
long-tail dilemma, i.e., fail to generalize on new scenarios, unexpected events, or traffic patterns~\cite{teng2023motion,chen2024end}. Besides, these approaches typically lack interpretability, making it difficult to explain their actions and hindering trust and regulatory approval.

To address these problems, several recent works explore the potential of VLMs for AD. 
LingoQA~\cite{marcu2312lingoqa} and Dolphins~\cite{ma2023dolphins}, for example, employ VQA to bridge the gap between data-driven driving and user trust. Besides, decision-making and planning are also being integrated into VLMs, as seen in DriveVLM~\cite{tian2024drivevlm}, DriveLM~\cite{sima2023drivelm}, Reason2drive~\cite{nie2025reason2drive}, BEV-InMLMM~\cite{ding2024holistic}, OmniDrive~\cite{wang2024omnidrive}, where the training data is always divided into perception, prediction, and planning components. These models often produce outputs via a chain-of-thought (CoT) process, encompassing scene descriptions, action analysis, hierarchical planning, etc. Approaches such as DriveGPT4~\cite{xu2024drivegpt4}, VLP~\cite{pan2024vlp}, AsyncDriver~\cite{chen2025asynchronous} and LMDrive~\cite{shao2024lmdrive} attempt to directly map visual and linguistic inputs to planning or low-level control signals. The end-to-end AD systems based on VLMs offer strong interpretability, trustworthiness, and the ability to understand complex scenes.

\setlength{\tabcolsep}{2.5pt}
\begin{table}[!t]
\centering
\caption{Comparison between the existing datasets and our proposed dataset. V. and Cate. represent video and category.}
\vspace{-0.3cm}
\label{datasetcom}
\begin{tabular}{ccccc}
\hline
Dataset & Source & V. & QAs & Cate. \\
\hline
CODA-LM~\cite{li2024automated} & CODA~\cite{li2022coda}  &$\times$&1.5K& 3 \\
LingoQA~\cite{marcu2312lingoqa} & Self-collected &$\checkmark$ &1K& 4  \\
IDKB~\cite{lu2024can} & Internet &$\checkmark$ &20K& 4  \\
nuScenes-QA~\cite{qian2024nuscenes} & nuScenes~\cite{caesar2020nuscenes} &$\checkmark$&83K& 5 \\
DriveLM~\cite{sima2023drivelm} & nuScenes~\cite{caesar2020nuscenes}  &$\checkmark$&15K& 4  \\
DriveBench~\cite{sima2023drivelm} & nuScenes~\cite{caesar2020nuscenes}  &$\checkmark$&21K& 5  \\
MME-Realworld~\cite{zhang2024mme} & ~\cite{li2022coda,caesar2020nuscenes,sachdeva2024rank2tell}&$\times$ &5K& 15\\
NuInstruct~\cite{ding2024holistic} & nuScenes~\cite{caesar2020nuscenes}  &$\checkmark$&16K& 17 \\
\hline
\textbf{VLADBench} & \begin{tabular}[c]{@{}c@{}}~\cite{caesar2020nuscenes,mao2021once,malla2023drama,Argoverse2}\\~\cite{kim2019CVPRhad,li2022coda,guo2023visualrsk10k,Houben-IJCNN-2013gtsdb}\\~\cite{han2021soda10m,kotseruba2016joint,rasouli2019pie,marcu2312lingoqa}\end{tabular}  &$\checkmark$&12K& 29\\  
\hline
\end{tabular}
\vspace{-0.7cm}
\end{table}

\subsection{Benchmark and Metrics}
Established evaluation benchmarks like MME~\cite{fu2023mme}, Video-MME~\cite{fu2024video}, MMBench~\cite{liu2025mmbench} and Seed-Bench~\cite{li2024seed}, while valuable for foundation models, are not ideally suited for evaluating AD models, because that these benchmarks primarily comprise natural images, lacking the specific characteristics of driving scenarios, such as traffic elements and the dynamic interactions of participants. Recent works have introduced specified AD datasets with extensive VQA pairs. Nuscenes-QA~\cite{qian2024nuscenes}, CODA-LM~\cite{li2024automated}, VLAAD~\cite{park2024vlaad} and LingoQA~\cite{marcu2312lingoqa} 
start from scene description and analysis, general perception, action reasoning and driving suggestions. DriveLM~\cite{sima2023drivelm}, NuInstruct~\cite{ding2024holistic}, Reason2Drive~\cite{nie2025reason2drive} divide the data into perception, prediction, and planning with reasoning. DriveLM~\cite{sima2023drivelm} also includes behavior understanding, and NuInstruct~\cite{ding2024holistic} includes risk estimation. IDKB~\cite{lu2024can} mined plenty of questions about 4 traffic knowledge domains from various handbooks. DriveBench~\cite{xie2025vlms} further introduce corruption data for robustness validation. With the rapid advancement, a coarse categorization is insufficient to support a complete analysis of AD models. 

For the metric, language-based metrics like BLEU~\cite{papineni2002bleu}, ROUGE~\cite{lin2004rouge}, METEOR~\cite{banerjee2005meteor} and CIDEr~\cite{vedantam2015cider}, commonly used to evaluate question-answering models, however, demonstrate poor correlation with human judgment. This is problematic because semantically distinct sentences with opposite meanings also can receive similar scores, posing unacceptable risks in safety-critical AD applications. While recent metrics leveraging ChatGPT ratings~\cite{xu2024drivegpt4,li2024automated}, they exhibit positional and stylistic biases and produce inconsistent scores across iterations. In this paper, we revisit the simple yet effective metric: Accuracy. Through the close-form instruction annotations, we try to achieve a precise evaluation in terms of the no tolerance for evaluation error.

\begin{table}[!t]
\centering
\caption{Prompt setting of \textbf{VLADBench}. $^{*}$ denotes optional. }
\vspace{-0.3cm}
\label{tab:promt_setting}
\begin{tabular}{p{0.46\textwidth}}
\hline
\textbf{Most Question}\\
\hline
[Image / Video] [Visual Prompt]$^{*}$ [Question] [Tips]$^{*}$.\\
Select one as the answer from the list below:\\
\quad[Choice A, B, C, D, E]. \\ 
No explanation is needed. The best answer is:\\
\hline
\hline
\textbf{Other
Question: Detection, Traffic Graph, Trajectory}\\
\hline
[Image / Video] [Visual Prompt]$^{*}$ [Question] [Tips]$^{*}$.\\
\-[Output Format]. \\
No explanation is needed. The answer is:\\
\hline
\end{tabular}
\vspace{-0.7cm}
\end{table}

\section{Benchmark}
\label{sec:dataset}

\begin{figure*}[!t]
  \centering
   \includegraphics[width=\linewidth]{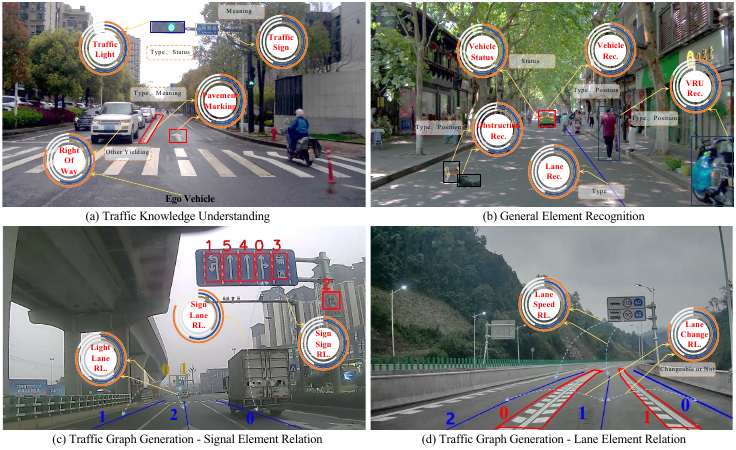}
    \vspace{-0.8cm}
    \caption{Real-world examples of the tasks in (a) Traffic Knowledge Understanding, (b) General Element Recognition, and (c, d) Traffic Graph Generation domains. `Rec.' and `RL' denote recognition and relation.
   }
   \label{fig:123}
\vspace{-0.6cm}
\end{figure*}

\subsection{Data Source and Annotation}
\noindent \textbf{Data Source}. A comprehensive and diverse benchmark dataset is able to reduce testing bias, which helps probe a thorough evaluation of zero-shot generalization capabilities and better expose the weakness of VLMs in various AD scenarios. As shown in ~\cref{datasetcom}, existing benchmarks often suffer from a lack of diversity. In contrast, our proposed dataset \textbf{VLADBench}, covering 5 domains, 11 aspects and 29 tasks, constructed from the 12 publicly available datasets: GTSDB~\cite{Houben-IJCNN-2013gtsdb}, JAAD~\cite{kotseruba2016joint}, PIE~\cite{rasouli2019pie}, HAD~\cite{kim2019CVPRhad}, nuScenes~\cite{caesar2020nuscenes}, SODA~\cite{han2021soda10m}, ONCE~\cite{mao2021once}, Argoverse2~\cite{Argoverse2}, CODA and CODA2022~\cite{li2022coda}, DRAMA~\cite{malla2023drama}, RS10K~\cite{guo2023visualrsk10k}, and LingoQA~\cite{marcu2312lingoqa}. The instance counts for the five domains TKU, GER, TGG, TAC, and EDP are 2369, 2812, 3090, 1303 and 2418, respectively. Detailed number for each task is included in the supplement.

\noindent \textbf{Annotation}. Based on the designed domains and tasks, we meticulously hand-select 2000 static and 3000 dynamic scenes for a diverse range of challenging driving situations. During the selection process, we control the visual prominence of objects and scenes to avoid immediate recognition. Existing datasets predominantly feature object- or caption-level annotations, lacking detailed and task-specific annotations. Consequently, we engage 5 human annotators for fine-grained annotation and implement a quality double-check with 2 professional researchers.  Each instance takes about 5 minutes to annotate.

\subsection{Instruction and Criterion}
\noindent \textbf{Instruction}. For most of the questions in the proposed dataset, we first construct each question-answer pair and then we collect all the answers in each task as a database. After that, we select the correct answer and randomly select the 
incorrect answers to form a choice list for each question. The choices in the list are semantically or structurally similar, increasing the ambiguity and difficulty of the question. The instruction format is listed in ~\cref{tab:promt_setting} and the length of the list ranges from 4 to 10. For the other types of questions, i.e., visual detection, traffic graph generation and trajectory planning, we specify the output format for each question to guide the instruction following. Additionally, some questions are constructed with visual prompts and descriptive tips. The visual prompts include the bounding boxes on the image or the coordinates of these boxes in the instructions, which are employed for regional perception representations. The tips consist of the perceptual descriptions within the scene, aiming at assisting the challenging task by providing accurate perceptual prior. These are particularly useful for domains like traffic graph generation.

\noindent \textbf{Criterion}.
For the evaluation of each task, the core metrics are accuracy and instruction compliance rate. Besides, there are IOU for detection in the recognition task, judgment accuracy for the intention judgment task, and L2 distance and collision rate for the ego trajectory planning task. The final score for each task is weighted by these metrics. Note that a rule-based filter is employed to align the responses generated by VLMs with the choice list in the instruction, removing symbols and special tokens.

\begin{figure*}[!t]
  \centering
   \includegraphics[width=\linewidth]{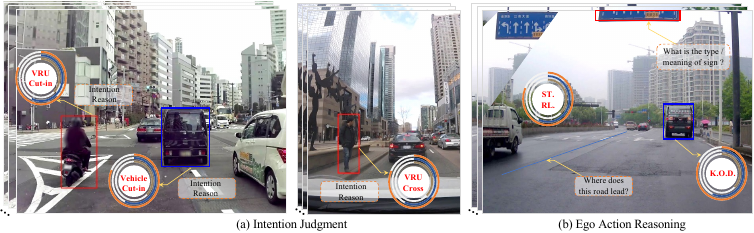}
  \vspace{-0.8cm}
   \caption{Examples in \textit{intention judgment} and \textit{ego action reasoning} aspects. ST.RL.: spatio-temporal reasoning, K.O.D.: key object detection.
   }
   \label{fig:45}
\vspace{-0.6cm}
\end{figure*}

\subsection{Data Statistics}
\label{sec:traffic_ele_rea}
\noindent \textbf{Traffic Knowledge Understanding}. This domain comprises two primary tasks: \textit{Road Traffic Signals} (encompassing the pavement marking, traffic sign, and traffic light tasks, with questions pertaining to type, status, meaning, and optical character recognition) and \textit{Road Passage Provisions} (Determining the right-of-way between ego vehicle and other traffic participants), as illustrated in ~\cref{fig:123} (a).

\noindent \textbf{General Element Recognition}. This domain consists of ~\textit{Background} and ~\textit{Foreground} elements. ~\textit{Background} includes light and weather conditions, while ~\textit{Foreground} focuses on lane recognition, vehicle recognition, vehicle status, vulnerable road user (VRU) recognition, and obstruction recognition. The recognition tasks also involve visual grounding questions, such as instructions with coordinates or object detection. The vehicle status task identifies the external states of the vehicle, such as brake lights, open doors, and trunk. The obstruction (including animals) recognition task, also addresses whether obstacles can be safely driven over, with examples illustrated in ~\cref{fig:123} (b).

\noindent \textbf{Traffic Graph Generation}. The aforementioned assessment allows for the perception of low-level scene elements, 
which forms the basis for a high-level understanding of the interrelationships between traffic elements, called traffic graph generation. This domain encompasses both \textit{Signal Element Relation}, with ~\cref{fig:123} (c) illustrating the light-lane relation, sign-lane relation and sign-sign relation, and \textit{Lane Element Relation}, with ~\cref{fig:123} (d) exhibiting lane speed relation and lane change relation. Based on the above-perceived results, we further organize descriptive tips for part questions, e.g., we provide the type and status of lights, and the type and meaning of the lanes for light-lane relation task to support more nuanced scene understanding.

\noindent \textbf{Target Attribute Comprehension.}
The comprehension of dynamic scenes is paramount in AD. In this domain, we address the temporal aspects by incorporating the prediction of future (unoccurred) events ~\textit{Intention Judgment} and the analysis of the past (occurred) events ~\textit{Behavior Understanding}. The visualization is presented in \cref{fig:45} (a) and \cref{fig:45} (b). ~\textit{Intention Judgment} is composed of vehicle cut-in, VRU cut-in, VRU cross, and long-short parking, with the questions pertaining to the underlying intention and motivation. ~\textit{Behavior Understanding} involves analyzing vehicle and VRU behavior by describing the sequence of temporal events. Moreover, we include pedestrian gesture analysis related to right-of-way determination in VRU behavior, further testing human-vehicle interaction capabilities. 

\noindent \textbf{Ego Decision-Making and Planning}. We construct this domain in a reasoning mechanism from \textit{Ego Action Reasoning}, the high-level \textit{Meta Decision-Making} and the final \textit{Ego Trajectory Planning}. \textit{Ego Action Reasoning} contains the fine-grained tasks: key object detection, drive efficiency, 
risk prediction, and spatio-temporal reasoning, all of which significantly influence subsequent driving strategies. Specifically, in spatio-temporal reasoning, we devise the challenge of inferring the state or meaning of part-occluded traffic signals, lane type, and lane destination at the end of a video sequence. At the moment of the last frame, information from previous frames, such as traffic signs and pavement markings, must be integrated and reasoned upon to answer these questions, as \cref{fig:45} (c) showcasing the real-world samples.
\textit{Meta Decision-Making} focuses on the short-term lateral and longitudinal decisions, which are tactical and on immediate execution. The decisions include but are not limited to straight, changing lane to the left/right, in-lane left/right avoidance, borrowing lane for left/right avoidance, accelerating, stop, maintaining, decelerating, and decelerating to stop.
\textit{Ego Trajectory Planning} is formulated as a vision and language task, given the critical perception and prediction results, along with high-level decisions. Besides, the ego status and the historical waypoints (last 2 seconds, given by four points) are included in the instruction. The VLMs then generate a feasible 3-second driving trajectory consisting of 6 waypoints.

\section{Experiments}
\label{sec:experiments}
\begin{table*}[!t]
\setlength{\extrarowheight}{3pt}
\setlength{\tabcolsep}{2pt}
\centering
\caption{Results evaluated on different VLMs. TKU: Traffic Knowledge Understanding, GER: General Element Recognition, TGG: Traffic Graph Generation, TAC: Target Attribute Comprehension, EDP: Ego Decision-Making and Planning. The gray, yellow and purple cell color denotes the open-source, closed-source, and domain-specific VLMs. The best score for each aspect in red. The detailed results about the 29 tasks are listed in the supplement. Note our baseline is excluded for comparison with existing models.}
\vspace{-0.3cm}
\label{all_results}
\resizebox{\textwidth}{!}{
\begin{tabular}{cccccccccccccccccccc||c}
 \hline
 &  & \cellcolor[HTML]{EFEFEF}IXC2.5 & \cellcolor[HTML]{EFEFEF}CV & \cellcolor[HTML]{EFEFEF}LoV & \cellcolor[HTML]{EFEFEF}QW & \cellcolor[HTML]{EFEFEF}IVL2 & \cellcolor[HTML]{EFEFEF}MCV & \cellcolor[HTML]{EFEFEF}IVL2 & \cellcolor[HTML]{EFEFEF}QW2 & \cellcolor[HTML]{EFEFEF}OV& \cellcolor[HTML]{EFEFEF}QW2.5 & \cellcolor[HTML]{EFEFEF}LV & \cellcolor[HTML]{FFFFC7}GEM & \cellcolor[HTML]{FFFFC7}GPT & \cellcolor[HTML]{CBCEFB}Senna& \cellcolor[HTML]{CBCEFB}Dols & \cellcolor[HTML]{CBCEFB}DriLM & \cellcolor[HTML]{CBCEFB}DriMM & \cellcolor[HTML]{CBCEFB}DriLM-B & \cellcolor[HTML]{CBCEFB}Ours \\
\multirow{-2}{*}{Domains} & \multirow{-2}{*}{Aspects} & 8B & 8B & 8B & 7B & 4B & 8B & 8B & 7B& 7B & 7B & 7B & 1.5pro & 4o & 7B & 9

B& 4B & 7B & 4B & 4B \\
 \hline
 & Road Traffic Signals & 24.89 & 47.00 & 37.46 & 47.40 & 48.97 & 43.91 & 54.97 & 54.77 & 56.89 &62.45& 57.49 & 67.56 & \textcolor{red}{69.09} &10.29& 28.39 & 55.04 & 57.15 & 52.56 & 65.65 \\
\multirow{-2}{*}{TKU} & Road Passage Provisions & 32.69 & 59.35 & 49.45 & 79.22 & 80.58 & 22.72 & 69.45 & 71.52 & 70.81 & 80.32 &74.11 & 42.98 & 78.96 &15.53& 21.10 & \textcolor{red}{81.36} & 42.33 & 73.85 & 80.58 \\
 \hline
 & Background & 22.54 & 58.79 & 63.93 & 63.97 & 66.83 & 70.76 & 70.49 & \textcolor{red}{71.07} & 69.11 & 69.29&69.11 & 65.71 & 68.35 &25.36& 58.75 & 64.46 & 70.31 & 68.21 & 71.61 \\
\multirow{-2}{*}{GER} & Foreground & 26.20 & 29.10 & 38.16 & 38.09 & 51.73 & 45.00 & 49.34 & 50.88 & 53.40 & 53.64&53.03 & 52.00 & 53.82 &15.51& 29.24 & 52.80 & \textcolor{red}{60.52} & 51.68 & 60.47 \\
 \hline
 & Signal Element Relation & 10.34 & 19.06 & 32.26 & 19.04 & 26.41 & 30.04 & 31.83 & 32.15 & 30.46 & 28.78&30.58 & 36.36 & \textcolor{red}{41.25} & 3.14&22.55 & 22.44 & 30.97 & 30.56 & 43.37 \\
\multirow{-2}{*}{TGG} & Lane Element Relation & 29.82 & 43.27 & 38.12 & 29.83 & 38.39 & 46.60 & 39.26 & 39.78 & 51.54 &40.14& 53.61 & \textcolor{red}{54.49} & 51.18 &41.22 & 26.29 & 19.22 & 21.59 & 44.06 & 44.48 \\
 \hline
 & Intention Judgment & 67.47 & 38.98 & 34.76 & \textcolor{red}{68.23} & 41.56 & 60.08 & 46.79 & 60.62 & 57.60 &53.42& 59.52 & 55.95 & 47.79 &52.79 &57.64 & 47.13 & 43.27 & 60.89 & 52.97 \\
\multirow{-2}{*}{TAC} & Behavior Understanding & 30.61 & 33.74 & 17.99 & 28.60 & 44.13 & 37.54 & 46.82 & 41.79 & 41.68 & 42.23&43.13 & 50.61 & \textcolor{red}{52.63} & 12.96& 0.11 & 40.78 & 40.11 & 42.91 & 42.91 \\
 \hline

 & Ego Action Reasoning & 45.91 & 36.89 & 52.03 & 64.24 & 46.47 & 58.85 & 61.91 & 54.93 & 47.35 &58.87& 56.50 & 61.20 & \textcolor{red}{65.75} &12.96& 58.72 & 55.95 & 52.99 & 56.60 & 69.73 \\
\multirow{-2}{*}{EDP} & Meta Decision-Making & 55.60 & 22.98 & 18.45 & 23.87 & 47.62 & 35.65 & 40.83 & 35.24 & 36.61 &35.48& 41.19 & \textcolor{red}{56.43} & 48.04 & 15.00&13.69 & 37.26 & 50.00 & 46.31 & 57.14 \\
 \hline
\multicolumn{2}{c}{Total} & 30.93 & 35.85 & 38.67 & 43.24 & 44.97 & 45.45 & 48.58 & 49.40 & 49.97 &50.75& 51.73 & 54.23 & \textbf{56.00} &20.21& 33.87 & 44.86 & 47.45 & 49.83 & 57.39\\
 \hline
\end{tabular}
}
\vspace{-0.7cm}
\end{table*}

\subsection{Baselines and Settings}
\noindent \textbf{Baselines}. To conduct a comprehensive evaluation on static and dynamic scenes, we compare 20 VLM models including the foundation and domain-specific models, which can be divided into the open-source VLMs: 
VILA-U (VU)~\cite{wu2024vilau}, InternLM-XComposer2.5 (IXC2.5)~\cite{zhang2023internlm}, Openflamingo~\cite{awadalla2023openflamingo}, CogVLM2 (CV)~\cite{hong2024cogvlm2}, LongVILA (LoV)~\cite{longvila}, QWen-VL (QW)\cite{bai2023qwen}, MiniCPM-V-2.6 (MCV)~\cite{yao2024minicpm}, InternVL2 (IVL2)~\cite{chen2024internvl}, Qwen2-VL (QW2)~\cite{wang2024qwen2}, OneVision (OV)~\cite{li2024llavaonevision}, LLaVA-Video (LV)~\cite{zhang2024llavanext-video}, QW2.5-VL (QW2.5)\cite{bai2025qwen2}, the closed-source VLMs:  Gemini-1.5-pro (GEM)~\cite{team2023gemini}, GPT-4o\footnote{https://openai.com/index/hello-gpt-4o/}, and the domain-specific VLMs: Dolphins (Dols)~\cite{ma2023dolphins}, Senna~\cite{jiang2024senna} (VLM part), DriveLM ~\cite{sima2023drivelm}(trained on DriveLM~\cite{sima2023drivelm} (DriLM) and trained on BDD~\cite{kim2018textual} (DriLM-B)), and DriveMM (DriMM)~\cite{huang2024drivemm}.


\noindent \textbf{Settings}. For the sequence data, we adjust the frames extraction to ensure all the frames are fed into the model. To ensure a fair comparison, system prompts are not utilized for models that offer them. As mentioned above, each task employs 2 to 3 metrics, which are weighted to compute the final score for the task. The instruction compliance rate is weighted at 0.2, while accuracy is weighted differently: 0.8 for most tasks, 0.5 for the TGG domain, and 0.7 for intention judgment tasks. The mean aspect score is then computed as an average of the task scores, with weights proportional to the number of tasks within each aspect. 

\noindent \textbf{Domain Data for Training}.
To further exploit the interactions among the 5 key domains for AD understanding, we start from a small-scale VLM, IVL2-4B ~\cite{chen2024internvl}, and train the DS models on individual domain datasets. These datasets, sourced from ~\cite{parikh2024idd,lu2024can,marcu2312lingoqa,sima2023drivelm,cao2024maplm,li2022coda,malla2023drama,kim2019CVPRhad,ma2023dolphins,wang2023openlane,guo2023visualrsk10k,xu2024drivegpt4,mao2023gpt}, contain a total of 1.4M QAs, covering perspectives from the ego vehicle, including single-view, sequential single-view, and multi-view. The type of each QA is classified using GPT-4. Besides, we also incorporate 1.3M QAs from general data for avoiding general ability loss. The IVL2-4B~\cite{chen2024internvl}, trained on 2.7M QAs, serves as our baseline in this paper. More details are provided in the supplement.

\subsection{Experimental Results}
First, we assess the existing open-source, closed-source, and DS models. The qualitative results across 10 aspects of \textbf{VLADBench} for the small-scale VLMs and closed-source VLMs, are listed in ~\cref{all_results}.
Besides, we present the results of large-scale VLMs in ~\cref{largescale_results} for a thorough assessment.
For comparison with the existing VLMs, we exclude our baseline model which serves for the following exploration.

Then, we conduct the domain experiments to explore the cognitive and reasoning interactions among the 5 key domains. The DS models trained on TKU data, GER data, TGG data, TAC data, EDP data, and the total data are compared with the base model for improvement visualization.

Finally, we briefly discuss how the understanding of the five key domains by AD-specialized VLMs will impact the final trajectory prediction.
\begin{figure*}[!ht]
  \centering
   \includegraphics[width=\linewidth]{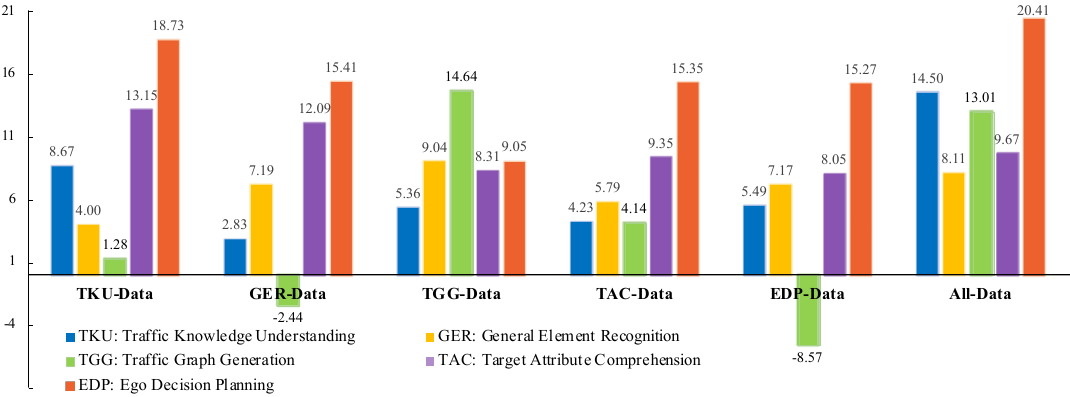}
  \vspace{-0.6cm}
   \caption{Gain chart of the five key domains. This chart shows the performance improvements of models trained on datasets categorized by the five key domains, evaluated on ADBench, compared to the base model.
   }
   \label{fig:gain}
\vspace{-0.5cm}
\end{figure*}

\subsubsection{Evaluation on \textbf{VLADBench}}

\setlength{\tabcolsep}{5.5pt}
\begin{table}[!t]
\centering
\caption{Results evaluated on Large-scale VLMs.}
\label{largescale_results}
\vspace{-0.3cm}
\resizebox{0.45\textwidth}{!}{
\begin{tabular}{ccccccc}
\hline
\multirow{2}{*}{Aspects} & VU    & OV    & LV    & IVL2  & QW2 & QW2.5  \\
                         & 40B   & 72B   & 72B   & 76B   & 72B  & 72B  \\
\hline
                         
Road Traffic Signals     & 32.32 & 54.15 & 59.09 & 59.94 & 68.31&\textcolor{red}{70.76} \\
Road Passage Provisions  & 46.34 & 76.96 & 80.06 & 71.07 & \textcolor{red}{81.36} &80.58\\
\hline
Background               & 65.13 & 70.67 & 71.43 & \textcolor{red}{72.68} & 70.00 &71.96\\
Foreground               & 42.37 & 51.91 & 52.27 & 56.36 & \textcolor{red}{60.56} &56.51\\
\hline
Signal Element Relation  & 32.06 & 30.63 & 31.72 & 33.71 & 29.91 &\textcolor{red}{40.14}\\
Lane Element Relation    & 39.40 & 51.58 & \textcolor{red}{55.69} & 40.32 & 49.95 &54.79\\
\hline
Intention Judgment       & 35.57 & 50.30 & 50.43 & 53.57 & \textcolor{red}{60.66} &54.26\\
Behavior Understanding   & 11.5  & 41.56 & 44.36 & 47.49 & 45.47 &\textcolor{red}{50.28}\\
\hline
Ego Action Reasoning     & 52.84 & 56.26 & 57.13 & 65.72 & 64.49 &\textcolor{red}{67.15}\\
Meta Decision-Making     & 19.29 & 53.45 & 38.45 & \textcolor{red}{58.33} & 53.81 &47.86\\
\hline
Total                    & 38.76 & 50.21 & 51.62 & 53.20 & 56.51 &\textbf{58.00}\\
\hline
\end{tabular}}
\vspace{-0.7cm}
\end{table}

\noindent \textbf{Holistic Results}. 
The top score is held by the large-scale QW2.5-72B~\cite{wang2024qwen2}, which achieves 58.00\%, and followed by GPT-4o. For small-scale VLM models, LV~\cite{zhang2024llavanext-video} leads with a score of 51.73\%, which is 6.27\% below the maximum. Even state-of-the-art VLMs such as GPT-4o and QW2.5~\cite{wang2024qwen2} achieve less than 60\% accuracy on our proposed benchmark, demonstrating the significant gap between current VLMs and human-level capabilities in real-world driving scenarios.

\noindent \textbf{Granular Results}. Through the results on 10 secondary aspects, the main findings are as follows: 
\begin{itemize}
    \item In TKU, \textit{Road Traffic Signal} represents a fundamental knowledge of AD. Existing open-source models, with the exception of large-scale QW series~\cite{wang2024qwen2,bai2025qwen2}, still remain a large room for performance improvement.
    \item On \textit{Signal Element Relation} aspect, GPT-4o and  QW2.5-72B~\cite{bai2025qwen2} showcase the superiority, where it outperforms the second best model by nearly 20\%, demonstrating the excellent spatial reasoning capability for AD scenarios.
    \item On \textit{Intention Judgment} aspect, almost all the models exhibit unsatisfactory performance. However, QW-7B
    \cite{bai2023qwen}, trained without sequential data, achieves promising results, suggesting that VLMs can predict potential sequential events through training on non-sequential events.
\end{itemize}
\noindent \textbf{Large-scale vs. Small-scale VLMs}. 
To investigate how the performance of a general model varies with scalability, we present 6 large-scale models in \cref{largescale_results}. Firstly, the large-scale VLM models do not always surpass the small-scale models. Among the large-scale models, only half outperform the best small-scale VLM model LV-7B~\cite{zhang2024llavanext-video}. Secondly, for a given model, scaling up the language model typically results in a performance improvement. However, this trend is not universally consistent in AD. For instance, with LV\cite{zhang2024llavanext-video}, the larger model generally outperforms its smaller counterpart across most aspects, yet it underperforms in the important ~\textit{intention judgment} and ~\textit{meta decision-making} aspects, resulting in a lower final score.

    

\noindent \textbf{Domain-specific Results}.
As training data from the AD domain is incorporated, the DS model exhibits outstanding performance in certain tasks, e.g., Dols\cite{ma2023dolphins} on vehicle cut-in task (86.70 ranked 1st), DriLM\cite{sima2023drivelm} on traffic light task (75.14 ranked 1st), DriMM\cite{huang2024drivemm} on foreground aspect (60.52 ranked 2nd). However, these DS models always perform poorly in TAC domain. By comparing DriLM\cite{sima2023drivelm} and DriLM-B\cite{sima2023drivelm} with the base model IVL2-4B~\cite{chen2024internvl}, we can further observe that biased domain-specific data will lead to a loss of generalization ability in unseen tasks. DriLM\cite{sima2023drivelm} outperforms the base model in fundamental traffic knowledge but performs significantly worse in the TAC domain. Meanwhile, DriLM-B\cite{sima2023drivelm} excels in \textit{intention Judgment} but falls short in \textit{meta decision-making} compared to the base model. DS Data bias will influences the model capabilities, and single-direction optimization may lead to a loss of generalization in other tasks even within the same domain.

\subsubsection{Interactions in Key Domains of \textbf{VLADBench}}
As discussed above, biased domain-specific training data can enhance the performance in certain specialized areas of autonomous driving but may loss the generalization ability in tasks that require broader and more general knowledge. To deeply explore the interrelationships among the 5 key domains, we train DS models using different DS datasets (with generic data kept constant) and test them on \textbf{VLADBench}. The gain chart, comparing these models to the base model, is shown in ~\cref{fig:gain}. It can be concluded that: 
\begin{itemize}
    \item The role of each domain data is not isolated, it also positively influences other domains. For example, TKU data boosts the EDP domain significantly, GER data benefits TAC domain (especially \textit{intention judgment}), TGG data enhance the understanding of the traffic element (lanes and traffic signs).
    \item Synergy effect occurs when combining all the datasets for training. When all the data trained together, the performance across all domains (except for TGG) is higher than when training each domain individually, e.g., TKU performance increased by 67\%.
    \item Although the all-data model achieved significant improvements in the TGG domain, models trained on GER and EDP data both experienced negative gains in the TGG domain, suggesting that the collected GER and EDP training datasets are still basised for TGG domain.

\setlength{\tabcolsep}{5.2pt}
\begin{table*}[!ht]
\caption{Motion planning performance. The metrics are following the setting in UniAD~\cite{hu2023planning} and ST-P3~\cite{hu2022stp3}.}
\label{tra_domain}
\vspace{-0.2cm}
\begin{tabular}{c|cccccccc|cccccccc}
\hline
\multirow{3}{*}{Models} &
  \multicolumn{8}{c|}{ST-P3} &
  \multicolumn{8}{c}{UniAD} \\ \cline{2-17} 
 &
  \multicolumn{4}{c|}{L2(m)$\downarrow$} &
  \multicolumn{4}{c|}{Collision(\%)$\downarrow$} &
  \multicolumn{4}{c|}{L2(m)$\downarrow$} &
  \multicolumn{4}{c}{Collision(\%)$\downarrow$} \\ \cline{2-17} 
 &
  1s &
  2s &
  3s &
  \multicolumn{1}{c|}{Avg.} &
  1s &
  2s &
  3s &
  Avg. &
  1s &
  2s &
  3s &
  \multicolumn{1}{c|}{Avg.} &
  1s &
  2s &
  3s &
  Avg. \\ \hline
ST-P3 &
  1.28 &
  2.03 &
  2.81 &
  \multicolumn{1}{c|}{2.04} &
  0.14 &
  0.72 &
  1.28 &
  0.71 &
  - &
  - &
  - &
  \multicolumn{1}{c|}{-} &
  - &
  - &
  - &
  - \\
UniAD &
  - &
  - &
  - &
  \multicolumn{1}{c|}{-} &
  - &
  - &
  - &
  - &
  0.47 &
  1.80 &
  3.73 &
  \multicolumn{1}{c|}{3.00} &
  0.13 &
  0.53 &
  1.50 &
  0.72 \\
  \hline
TKU &
  0.30 &
  0.67 &
  1.15 &
  \multicolumn{1}{c|}{0.70} &
  0.13 &
  0.25 &
  0.82 &
  0.40 &
  0.43 &
  1.28 &
  2.41 &
  \multicolumn{1}{c|}{1.37} &
  0.13 &
  0.51 &
  2.53 &
  1.06 \\
GER &
  0.28 &
  0.61 &
  1.04 &
  \multicolumn{1}{c|}{0.65} &
  0.06 &
  0.26 &
  0.69 &
  0.34 &
  0.41 &
  1.16 &
  2.16 &
  \multicolumn{1}{c|}{1.24} &
  0.13 &
  0.52 &
  2.07 &
  0.91 \\
TGG &
  0.34 &
  0.78 &
  1.31 &
  \multicolumn{1}{c|}{0.81} &
  0.00 &
  0.38 &
  0.95 &
  0.44 &
  0.50 &
  1.50 &
  2.66 &
  \multicolumn{1}{c|}{1.55} &
  0.00 &
  0.88 &
  2.40 &
  1.10 \\
TAC &
  0.35 &
  0.83 &
  1.43 &
  \multicolumn{1}{c|}{0.87} &
  0.00 &
  0.28 &
  1.09 &
  0.46 &
  0.53 &
  1.62 &
  2.94 &
  \multicolumn{1}{c|}{1.70} &
  0.00 &
  0.76 &
  3.40 &
  1.39 \\
EDP &
  0.28 &
  0.71 &
  1.25 &
  \multicolumn{1}{c|}{0.75} &
  0.13 &
  0.25 &
  0.77 &
  0.38 &
  0.41 &
  1.42 &
  2.66 &
  \multicolumn{1}{c|}{1.50} &
  0.13 &
  0.50 &
  2.51 &
  1.05 \\
All &
  0.27 &
  0.60 &
  1.04 &
  \multicolumn{1}{c|}{0.64} &
  0.13 &
  0.25 &
  0.61 &
  0.33 &
  0.39 &
  1.15 &
  2.15 &
  \multicolumn{1}{c|}{1.23} &
  0.13 &
  0.51 &
  1.65 &
  0.76 \\
  \hline
\end{tabular}
\vspace{-0.5cm}
\end{table*}

    \item Adjusting the data ratio between different domains may lead to better training results. For example, both TKU data and GER data contribute more to TAC than TAC data, indicating that selecting an appropriate ratio could yield improved performance. 
\end{itemize}

\subsubsection{Contribution of Key Domains for Motion Planning}
In \textbf{VLADBench}, tasks for comprehension dominate the evaluation. After assessing the 5 key domains for AD understanding tasks, we finally evaluate the contributions of 5 key domains for trajectory prediction. Note that the goal is not to pursue the state-of-the-art results. For training, we construct about 4K dataset from nuScenes~\cite{caesar2020nuscenes}, which includes scene analysis and trajectory points, and then train the model by incorporating individual domain-specific data. 
The quantitative motion planning results are shown in \cref{tra_domain}. It can be observed that the GER is the most important domain for trajectory prediction, followed by EDP domain. The results from TKU domain are comparable to those from EDP domain, suggesting that the understanding of traffic knowledge plays a crucial role, which is a capability that traditional models are unable to achieve. Although the experimental results from TGG and TAC domains perform poorly in terms of L2 distance, they significantly reduce the collision rate in the short term. More details about the trajectory dataset and results from open-source and domain-specific models are presented in the supplement materials. 
\subsection{Further Analysis}

\noindent \textbf{Bottlenecks in Traffic Graph Generation}.
In TGG domain, analyzing the relations between elements presents significant challenges. As discussed above in \cref{sec:traffic_ele_rea}, we further incorporate additional descriptive guidance about the traffic elements as tips within the instructions. Experiments after and before adding these tips showcase a 10\% improvement in the light-lane relation task and a 20\% improvement in the lane change relation task, suggesting that embedding traffic-related knowledge can directly enhance knowledge graph construction. Nevertheless the accuracy remains at approximately 60\%, indicating that spatial reasoning ability is still limited. The detailed experimental results are listed in the supplementary material.
 
\noindent \textbf{The larger, the better?}
OV-7B \cite{li2024llavaonevision} and LV-7B \cite{zhang2024llavanext-video} perform as the top models at the small scale. However, when the language model is scaled up to 72B, the vision encoder, SigLIP \cite{siglip}, remains unchanged, and the observed superiority no longer holds. Specifically, OV-72B \cite{li2024llavaonevision} shows only marginal improvement, and LV-72B even \cite{zhang2024llavanext-video} experiences a performance decline. In contrast, IVL2-76B\cite{zhang2024llavanext-video}, with a significantly larger vision encoder (scaled from 300M to 6B parameters), achieved first place across two aspects. QW2-72B\cite{wang2024qwen2} and QW2.5-72B\cite{bai2025qwen2}, featuring a larger vision encoder than OV\cite{li2024llavaonevision} and LV\cite{zhang2024llavanext-video} and employing a dynamic resolution mechanism to avoid visual information loss, approaches the performance of closed-source models, achieving a well-balanced performance across cognitive and reasoning tasks. These findings suggest that a large or specialized vision encoder may be more critical than merely scaling up the language model for AD.



\section{Conclusion and Limitation}
\label{sec:con_lim}
\noindent \textbf{Conclusion}. 
In this paper, we present a fine-grained benchmark for evaluating large vision-language models on autonomous driving. The proposed \textbf{VLADBench} covers 5 key domains, 11 aspects and 29 tasks, addressing critical gaps in current datasets, including coarse-grained categories, and a lack of dynamic element analysis and diversity. Extensive experiments on general and domain-specific models uncover the significant performance gaps across a wide range of tasks. Our in-depth experiments further reveals the interactions among the five key domains, and the individual contribution for motion planning performance.



\noindent \textbf{Limitations}. There are still several limitations: 1) The current benchmark focuses on evaluating the understanding and reasoning capabilities from the perspective view. Future research will incorporate multi-view inputs to further assess the 3D spatial perception capabilities of these models. 2) The training of domain-specific models in this paper is straightforward. Exploring the scalability of domain-specific models and optimizing data sampling strategies are the crucial directions for future researches.
{
    \small
    \bibliographystyle{ieeenat_fullname}
    \bibliography{main}
}
\clearpage
\setcounter{page}{1}
\maketitlesupplementary

In the supplementary, we present more details about the benchmark, training setting, and results on each tasks. Then we conduct additional experimental results related to the tips in instructions for ~\textit{Traffic Graph Generation} aspect.
Also, we compare the domain-specific model Dols~\cite{ma2023dolphins} and the underlying foundation model Openflamingo~\cite{awadalla2023openflamingo} on static tasks. Finally, we provide a detailed definition of each task in the proposed dataset \textbf{VLADBench}, illustrated with examples, including the images and corresponding questions. Note that the visual prompts and the choice lists are omitted.

\section{Benchmark details}
Following the selection principle, we first choose scenarios, annotate the visual elements with descriptions, design the questions, and then annotate the correct answers. 5 human annotators perform fine-grained annotations, with 2 researchers verifying the results. Each instance takes about 5 to annotate. The instance counts for the five domains TKU, GER, TGG, TAC, and EDP are 2369, 2812, 3090, 1303, and 2418. The instance numbers of Traffic Light, Pavement Marking, Traffic Sign, Right Of Way, VRU Recognition, Vehicle Recognition, Vehicle Status, Lane Recognition, Obstruction Recognition, Light, Weather, Sign-Sign Relation, Light-Lane Relation, Sign-Lane Relation, Lane-Change Relation, Road-Speed Relation, VRU Cut-in, VRU Cross, Vehicle Cut-in, Long-Short Parking, VRU Bahavior, Vehicle Bahavior, Key Obstruction Detection, Risk Prediction, Drive Efficiency, Spatio-Temporal Reasoning, Lateral, Longitudinal, and Trajectory tasks are 795, 564, 701, 309, 424, 223, 257, 780, 680, 200, 248, 192, 702, 1072, 784, 340, 267, 276, 261, 320, 99, 80, 547, 272, 303, 161, 235, 101, and 799, respectively.

\section{Training Setting}
\noindent \textbf{Training Data.}
The domain-specific dataset are collected from ~\cite{parikh2024idd,lu2024can,marcu2312lingoqa,sima2023drivelm,cao2024maplm,chen2024automated,malla2023drama,kim2019CVPRhad,ma2023dolphins,wang2023openlane,guo2023visualrsk10k,xu2024drivegpt4,mao2023gpt}, including DriveLM-nuscenes~\cite{sima2023drivelm} (377956 QAs), LingoQA~\cite{marcu2312lingoqa} (413829 QAs), CODA-LM ~\cite{chen2024automated} (20495 QAs), Dolphins~\cite{ma2023dolphins} (102025 QAs), IDKB~\cite{lu2024can} (188486 QAs), MapLM~\cite{cao2024maplm} (143252 QAs), DriveGPT4~\cite{xu2024drivegpt4} (26751 QAs). Besides, we employ structured rules to generate 109309 QAs, using the original annotations from ~\cite{parikh2024idd,guo2023visualrsk10k,kim2019CVPRhad,malla2023drama}. Then we use GPT-4o to increase the diversity of the 109309 QAs. 

For the trajectory training data, we selected 4,072 samples from nuScenes~\cite{caesar2020nuscenes}. To generate analytical data for each scenario, we first collect questions related to each sample from ~\cite{sima2023drivelm,ding2024holistic}. Then, we utilize GPT-4o to summarize these questions and transform them into declarative statements. Finally, by incorporating images, we use GPT-4o to generate detailed scenario analyses. We will publicly release this dataset to support research in the autonomous driving field.

\noindent \textbf{Training Details.} The training framework is inherited from IVL2~\cite{chen2024internvl}. We finetune the pre-trained IVL2-4B with full parameters (including the vision encoder) for 2 epochs, with a batch size of 1 and a learning rate of 1e{-5}. The max token length is set to 4096. All experiments are conducted on 16 nodes, each equipped with 8 V100 GPUs, with each task taking approximately 24 hours.
\section{Bottlenecks in Traffic Graph Generation}

As discussed, \textit{traffic graph generation} is challenging, even with the provision of visual prompts. To dig into underlying reasons further, we incorporate descriptive guidance on traffic knowledge, including the meaning of signs, types of lights, and lane characteristics. \cref{sup_tips} and \cref{tip_ds_models} list the accuracy improvement across the five tasks. Almost all models show improvement, suggesting that embedding traffic-related knowledge aids in graph construction. For example, GPT-4o demonstrates a 10.63\% improvement in sign-lane relation and a 17.21\% improvement in lane speed relation tasks. However, accuracy remains around 60\%, highlighting ongoing limitations in relational reasoning. The marginal improvement in the sign-sign relation task further underscores these constraints. 
Put things together, two points can be drawn: 1) Current Visual Language Models (VLMs) still lack sufficient perception and understanding of traffic knowledge. 2) Even when provided with all the relevant knowledge, existing VLMs still exhibit weak reasoning capabilities in traffic graph tasks.

\begin{table*}[t!]
\centering
\caption{The improvement rate of accuracy on $\textit{traffic graph generation}$ aspect. S.S. RL.: Sign-Sign Relation, S.L.RL.: Sign-Lane Relation, L.L.RL.: Light-Lane Relation, L.S. RL.: Lane Speed Relation, L.C. RL.: Lane Change Relation.}
\vspace{-0.2cm}
\label{sup_tips}
\resizebox{\textwidth}{!}{
\begin{tabular}{ccccccccccccccccccccc}
\hline
 & \multicolumn{1}{l}{VU} & IXC2.5 & VILA & CV & LoV & QW & MCV & IVL2 & QW2 & OV & QW2.5 & LV  &VILA & OV & LV & IVL2 & QW2 & QW2.5 & \cellcolor[HTML]{FFFFC7}GEM & \cellcolor[HTML]{FFFFC7}GPT \\
\multirow{-2}{*}{Task} & 8B & 8B & 8B & 8B & 8B & 7B & 8B & 8B & 7B & 7B & 7B & 7B  & 40B & 72B & 72B & 72B & 72B & 72B & \cellcolor[HTML]{FFFFC7}1.5pro & \cellcolor[HTML]{FFFFC7}4o \\
\hline

S.S.RL & 0.52 & 0.26 & 0.78 & 0.00 & 0.26 & 0.26 & 3.65 & 0.78 & 0.00 & 0.00 &0.26& 0.26  & 0.78 & 1.04 & 1.04 & 0.00 & 0.00 &0.78& 3.13 & 5.21 \\
S.L.RL & 5.60 & 5.50 & 6.16 & 6.86 & 6.25 & 4.99 & 5.13 & 5.60 & 6.16 & 5.18 & 3.78&4.62  & 6.39 & 3.87 & 4.94 & 5.60 & 5.32 & 10.03&9.14 & 10.63 \\
\multicolumn{1}{l}{L.L.RL} & 6.48 & 4.99 & 2.92 & 11.11 & 5.27 & 6.62 & 6.41 & 7.55 & 5.48 & 5.27&5.91 & 2.99  & 3.92 & 12.61 & 13.11 & 8.05 & 8.90 & 8.55&10.75 & 8.19 \\
\rowcolor[HTML]{EFEFEF} 
MEAN & 5.42 & 4.81 & 4.48 & 7.71 & 5.32 & 5.11 & 5.44 & 5.82 & 5.32 & 4.70 & 4.20&3.61  & 4.96 & 6.71 & 7.48 & 5.93 & 6.08 & 8.60&9.13 & \textbf{9.23} \\
\hline

L.S.RL & 4.56 & 14.26 & 4.85 & 6.18 & 2.06 & 0.29 & 8.97 & 9.12 & 12.65 & 8.24 & 14.26&8.97  & 7.94 & 12.06 & 12.06 & 15.88 & 17.79 &17.06& 12.94 & 17.21 \\
L.C.RL & 0.00 & 4.21 & 7.97 & 14.54 & 7.91 & 11.35 & 13.84 & 8.35 & 7.72 & 17.22 & 7.78&17.28  & 7.97 & 19.32 & 17.98 & 7.72 & 13.39 & 18.88&19.77 & 15.69 \\
\rowcolor[HTML]{EFEFEF} 
MEAN & 1.38 & 7.25 & 7.03 & 12.01 & 6.14 & 8.01 & 12.37 & 8.59 & 9.21 & 14.50 & 9.74&14.77  & 7.96 & 17.13 & 16.19 & 10.19 & 14.72 & \textbf{18.33}&17.70 & 16.15 \\
\hline
\end{tabular}
}
\end{table*}

\setlength{\tabcolsep}{0.8pt}
\begin{table}[!t]
\centering
\caption{The improvement rate of accuracy on traffic graph generation aspect. The results are from the domain-specific models.}
\vspace{-0.2cm}
\label{tip_ds_models}
\resizebox{0.45\textwidth}{!}{
\begin{tabular}{c|ccccccc}
\hline
\multirow{2}{*}{Task} &Dols& Senna & DriLM & DriMM & DriLM-B & IVL2&Ours  \\
                      & 9B & 7B    & 4B    & 7B    & 4B     & 4B  & 4B    \\
\hline
S.S.RL.               &0.78& 0.00  & 0.78  & 1.82  & 0.00    & 0.00&0.52  \\
S.L.RL.               &4.20& 0.42  & 1.40  & 6.86  & 4.24    & 2.85& 7.00  \\
L.L.RL.               & 4.34&0.07  & 9.33  & 4.70  & 7.34    & 3.99&8.69  \\
\rowcolor[HTML]{EFEFEF}
MEAN                  & 3.92&0.25  & 4.17  & 5.60  & 4.93    & 2.98&\textbf{6.97}  \\
\hline
L.S.RL.               & 0.00&0.00  & 7.21  & 7.50  & 8.38    & 9.12&7.79  \\
L.C.RL.               & 7.72&17.28 & 0.00  & 0.00  & 7.72    &7.79& 11.80 \\
\rowcolor[HTML]{EFEFEF} 
MEAN                  & 5.38&\textbf{12.06} & 2.18  & 2.27  & 7.92    & 8.14&10.59\\
\hline
\end{tabular}}
\end{table}


\begin{table*}[t!]
\centering
\caption{Results on the static parts of our proposed dataset from the domain-specific model (Dols) and its foundation model (Openflamingo). P.M.: Pavement Marking, T.S.: Traffic Sign, T.L.: Traffic Light, LI.: Light, WE.: Weather, L.Rec.: Lane Recognition, V.S.: Vehicle Status, V.Rec.: Vehicle Recognition, VRU.Rec.: VRU Recognition, O.Rec.: Obstruction Recognition, S.S. RL.: Sign-Sign Relation, S.L.RL.: Sign-Lane Relation, L.L.RL.: Light-Lane Relation, L.S. RL.: Lane Speed Relation, L.C. RL.: Lane Change Relation, K.O.D: Key Object Detection.}
\vspace{-0.2cm}
\label{sup_DOLS}
\resizebox{\textwidth}{!}{
\begin{tabular}{cccccccccccc|ccccc}
\hline
Task & P.M. & T.S. & T.L. & LI. & WE. & L.Rec. & V.S. & V.Rec. & VRU.Rec & O.Rec. & S.S.RL. & S.L.RL. & L.L.RL. & L.S.RL & L.C.RL & K.O.D. \\
\hline
Openflamingo & 30.92 & 31.61 & 58.04 & 45.2 & 54.84 & 46.21 & 27.32 & 0.00 & 0.38 & 0.00 & 1.04 & 2.10 & 5.84 & 1.18 & 7.46 & 0.00 \\
Dolphins & 12.09 & 25.02 & 42.92 & 56.40 & 60.65 & 30.97 & 3.97 & 0.00 & 0.00 & 0.00 & 0.78 & 4.20 & 4.34 & 0.00 & 7.72 & 74.11\\
\hline
\end{tabular}}
\end{table*}

\setlength{\tabcolsep}{2.5pt}
\begin{table*}[!t]
\caption{More motion planning result from the State-of-the-art models and DS models.}
\vspace{-0.3cm}
\label{more_tra}
\centering
\begin{tabular}{c|cccccccc|cccccccc}
\hline
\multirow{3}{*}{Data} & \multicolumn{8}{c|}{ST-P3}                                                          & \multicolumn{8}{c}{UniAD}                                                          \\ \cline{2-17} 
                      & \multicolumn{4}{c|}{L2(m)}                         & \multicolumn{4}{c|}{Collision} & \multicolumn{4}{c|}{L2(m)}                         & \multicolumn{4}{c}{Collision} \\ \cline{2-17} 
                      & 1s    & 2s    & 3s    & \multicolumn{1}{c|}{Avg}   & 1s    & 2s     & 3s     & Avg  & 1s    & 2s    & 3s    & \multicolumn{1}{c|}{Avg}   & 1s    & 2s    & 3s    & Avg   \\ \hline
ST-P3                    & 1.28  & 2.03  & 2.81  & \multicolumn{1}{c|}{2.04}  & 0.14  & 0.72   & 1.28   & 0.71 & -     & -     & -     & \multicolumn{1}{c|}{-}     & -     & -     & -     & -     \\
Uniad                 & -     & -     & -     & \multicolumn{1}{c|}{-}     & -     & -      & -      & -    & 0.47  & 1.80  & 3.73  & \multicolumn{1}{c|}{3.00}  & 0.13  & 0.53  & 1.50  & 0.72  \\
\hline
IVL-4B                & 5.93  & 7.39  & 8.91  & \multicolumn{1}{c|}{7.41}  & 6.51  & 8.49   & 9.73   & 8.25 & 6.92  & 9.58  & 12.66 & \multicolumn{1}{c|}{9.72}  & 7.66  & 10.44 & 14.00 & 10.70 \\
DriveMM               & 11.46 & 15.05 & 18.74 & \multicolumn{1}{c|}{15.08} & 1.92  & 6.46   & 12.93  & 7.12 & 10.24 & 12.17 & 14.06 & \multicolumn{1}{c|}{12.06} & 1.06  & 2.94  & 5.69  & 3.23  \\
GPT-4o                & 4.74  & 8.41  & 11.74 & \multicolumn{1}{c|}{8.30}  & 4.26  & 10.98  & 11.11  & 8.79 & 3.64  & 5.57  & 7.36  & \multicolumn{1}{c|}{5.53}  & 2.78  & 6.17  & 7.92  & 5.62  \\
Gemini-1.5pro                   & 3.70  & 6.67  & 9.44  & \multicolumn{1}{c|}{6.61}  & 5.11  & 8.68   & 7.54   & 7.11 & 2.78  & 4.35  & 5.84  & \multicolumn{1}{c|}{4.33}  & 3.32  & 5.62  & 6.45  & 5.13 \\
\hline
\end{tabular}
\end{table*}

\section{Dols versus Base Model}
Dols~\cite{ma2023dolphins} focuses on autonomous vehicle behavior understanding and achieves state-of-the-art performance in the vehicle cut-in intention judgment task. However, Dols~\cite{ma2023dolphins} perform quite worse on most other tasks. One reason for this is the poor performance of the underlying foundation model (Openflamingo~\cite{awadalla2023openflamingo}), as demonstrated by the results on the static parts of our proposed dataset in \cref{sup_DOLS}. Additionally, we observe a notable performance drop in the \textit{road traffic signals} aspect, with no improvements in recognition tasks or graph construction. These phenomenons highlight focusing on only one capability when adapting a foundational model to an autonomous driving domain is suboptimal. This approach prevents other relevant autonomous driving capabilities from improving and may even lead to significant performance degradation.

\begin{table*}[!ht]
\caption{ Detailed results evaluated on different VLMs. For the abbreviation, P.M.: Pavement Marking, T.S.: Traffic Sign, T.L.: Traffic Light, R.O.W: Right Of Way, LI.: Light, WE.: Weather, L.Rec.: Lane Recognition, V.S.: Vehicle Status, V.Rec.: Vehicle Recognition, VRU.Rec.: VRU Recognition, O.Rec.: Obstruction Recognition, S.S. RL.: Sign-Sign Relation, S.L.RL.: Sign-Lane Relation, L.L.RL.: Light-Lane Relation, L.S. RL.: Lane Speed Relation, L.C. RL.: Lane Change Relation, VRU.CI.: VRU Cut-in, V.CI: Vehicle Cut-in, VRU.C: VRU Cross, L.S.P.: Long-Short Parking, V.B.: Vehicle Behavior, VRU. B: VRU Behavior, K.O.D: Key Object Detection, ST.R.: Spatio-temporal Reasoning, R.P: Risk Prediction, D.E.: Drive Efficiency, LO: Longitudinal, LA: Lateral.}
\label{detailed_results_each}
\resizebox{\textwidth}{!}{
\begin{tabular}{c|cccccccccccc|cccccccc|cccc||c}
\hline
\multirow{2}{*}{Task} & IXC2.5 & VILA  & CV    & LoV  & QW & IVL2& MCV   & IVL2  & QW2   & OV    & QW2.5 & LV    & VILA  & OV    & LV    & IVL2  & Q2-VL & GEM.   & GPT   & QW2.5 & Dols   & DriLM & DriMM & DriLM-B & Ours  \\ \cline{2-26} 
                      & 8B     & 8B    & 8B    & 8B   & 7B & 4B & 8B    & 8B    & 7B    & 7B    & 7B    & 7B    & 40B   & 72B   & 72B   & 76B   & 72B   & 1.5pro & 4o    & 72B   & 9B    & 4B    & 7B    & 4B      & 4B    \\ \hline
T.L.                  & 30.24  & 44.53 & 62.82 & 44.88 & 67.37&69.51&60.86 & 67.70 & 65.11 & 62.57 & 62.77 & 60.05 & 43.85 & 73.11 & 70.92 & 71.32 & 66.79 & 66.79  & 67.72 & 74.54 & 42.92 & \textcolor{red}{75.14} & 68.20 & 71.77   & 74.94 \\
P.M.                  & 22.59  & 31.21 & 38.51 & 50.60 & 36.42&38.62&34.72 & 46.38 & 57.38 & 58.37 & 62.91 & 63.51 & 31.03 & 53.72 & 65.11 & 51.38 & 66.88 & \textcolor{red}{73.01}  & 71.17 & 72.45 & 12.09 & 39.50 & 54.11 & 40.99   & 58.33 \\
T.S.                  & 20.68  & 13.50 & 35.89 & 18.46 & 33.58&34.01&32.10 & 47.45 & 40.94 & 49.27 & 61.71 & 49.73 & 20.29 & 32.98 & 40.83 & 53.92 & \textcolor{red}{71.18} & 64.05  & 68.96 & 65.11 & 25.02 & 44.74 & 47.05 & 40.09   & 61.00 \\
\rowcolor[HTML]{EFEFEF} 
MEAN                  & 24.89  & 30.32 & 47.00 & 37.46 & 47.40&48.97&43.91 & 54.97 & 54.77 & 56.89 & 62.45 & 57.49 & 32.32 & 54.15 & 59.09 & 59.94 & 68.31 & 67.56  & 69.09 & \textcolor{red}{70.76} & 28.39 & 55.04 & 57.15 & 52.56   & 65.65 \\
\hline
R.O.W.                & 32.69  & 46.34 & 59.35 & 49.45 & 79.22&80.58&22.72 & 69.45 & 71.52 & 70.81 & 80.32 & 74.11 & 46.34 & 76.96 & 80.06 & 71.07 & \textcolor{red}{81.36} & 42.98  & 78.96 & 80.58 & 21.10 & \textcolor{red}{81.36} & 42.33 & 73.85   & 80.58 \\
\rowcolor[HTML]{EFEFEF} 

MEAN                  & 32.69  & 46.34 & 59.35 & 49.45 & 79.22&80.58&22.72 & 69.45 & 71.52 & 70.81 & 80.32 & 74.11 & 46.34 & 76.96 & 80.06 & 71.07 & \textcolor{red}{81.36} & 42.98  & 78.96 & 80.58 & 21.10 & \textcolor{red}{81.36} & 42.33 & 73.85   & 80.58 \\
\hline
VRU.Rec               & 20.78  & 15.97 & 20.84 & 21.21 & 18.75&39.10&36.65 & 38.34 & 45.24 & 36.99 & 39.79 & 35.95 & 31.99 & 35.51 & 34.00 & 47.89 & \textcolor{red}{53.12} & 47.02  & 40.58 & 38.85 & 21.20 & 30.47 & 51.07 & 34.56   & 46.05 \\
V.Rec                 & 42.19  & 24.62 & 27.34 & 36.39 & 50.16&61.33&56.11 & 61.92 & 66.07 & 60.30 & 52.54 & 58.27 & 53.09 & 63.20 & 63.82 & \textcolor{red}{74.97} & 64.70 & 62.71  & 58.75 & 58.78 & 46.41 & 47.71 & 74.93 & 56.76   & 74.14 \\
V.S.                  & 23.66  & 26.77 & 24.67 & 41.95 &24.36 &51.05&54.09 & 49.26 & 47.16 & 54.09 & 52.61 & 52.53 & 48.40 & 53.62 & 54.94 & \textcolor{red}{59.69} & 54.16 & 55.41  & 54.86 & 55.95 & 3.97  & 51.36 & 55.49 & 48.79   & 56.96 \\
L.Rec                 & 32.92  & 45.97 & 54.36 & 49.26 &54.23 &57.72&47.90 & 48.62 & 52.51 & 67.69 & 63.49 & 66.15 & 49.85 & 67.54 & 64.05 & 64.44 & 62.87 & 51.26  & 63.49 & \textcolor{red}{70.87} & 30.97 & 73.44 & 66.49 & 66.62   & 74.26 \\
O.RG                  & 17.58  & 29.70 & 7.62  & 35.15 &32.87 &49.83&39.81 & 52.94 & 48.95 & 44.73 & 51.74 & 47.11 & 34.46 & 39.88 & 45.36 & 45.00 & \textcolor{red}{63.62} & 51.15  & 48.97 & 50.51 & 36.18 & 45.25 & 56.74 & 44.65   & 50.47 \\
\rowcolor[HTML]{EFEFEF} 

MEAN                  & 26.20  & 31.81 & 29.13 & 38.16 & 38.09&51.73&45.00 & 49.34 & 50.88 & 53.40 & 53.64 & 53.03 & 42.37 & 51.91 & 52.27 & 56.36 & \textcolor{red}{60.56} & 52.00  & 53.82 & 56.51 & 29.24 & 52.80 & 60.52 & 51.68   & 60.47 \\
\hline
LI.                   & 16.40  & 63.80 & 58.60 & 62.50 & 58.30&63.30&\textcolor{red}{70.40} & 66.30 & \textcolor{red}{70.40} & 69.60 & 68.40 & 70.00 & 62.20 & 70.00 & \textcolor{red}{70.40} & \textcolor{red}{70.40} & 67.20 & 56.40  & 67.20 & 73.20 & 56.40 & 63.60 & 68.70 & 66.40   & 68.40 \\
WE.                   & 27.50  & 65.81 & 58.95 & 65.08 & 68.55&69.68&71.05 & 73.87 & 71.61 & 68.71 & 70.00 & 68.39 & 67.50 & 71.21 & 72.26 & \textcolor{red}{74.52} & 72.26 & 73.23  & 69.27 & 70.97 & 60.65 & 65.16 & 71.61 & 69.68   & 74.19 \\
\rowcolor[HTML]{EFEFEF} 

MEAN                  & 22.54  & 64.91 & 58.79 & 63.93 &63.97 &66.83&70.76 & 70.49 & 71.07 & 69.11 & 69.29 & 69.11 & 65.13 & 70.67 & 71.43 & \textcolor{red}{72.68} & 70.00 & 65.71  & 68.35 & 71.96 & 58.75 & 64.46 & 70.31 & 68.21   & 71.61 \\
\hline
S.S.RL                & 0.25   & 21.09 & 10.00 & 20.78 & 16.93&19.90&26.41 & 21.02 & 20.00 & 22.00 & 20.73 & 22.99 & 20.57 & 21.26 & 22.96 & 20.00 & 20.13 & 25.36  & \textcolor{red}{31.87} & 21.45 & 17.58 & 9.01  & 23.20 & 20.00   & 19.24 \\
L.R.RL                & 10.84  & 32.07 & 22.17 & 33.02 &19.59 &29.14&31.58 & 34.77 & 33.44 & 32.61 & 31.85 & 32.14 & 31.86 & 39.45 & 38.75 & 39.81 & 33.82 & 41.12  & 42.90 & \textcolor{red}{43.65} & 23.73 & 28.09 & 28.49 & 34.55   & 49.19 \\
S.L.RL                & 11.82  & 33.48 & 18.64 & 33.81 &19.05&25.79 &29.68 & 31.84 & 33.48 & 30.57 & 28.21 & 30.93 & 34.25 & 26.54 & 28.69 & 32.17 & 29.09 & 35.21  & \textcolor{red}{41.84} & 41.19 & 22.67 & 21.14 & 33.99 & 29.84   & 40.20 \\
\rowcolor[HTML]{EFEFEF} 

MEAN                  & 10.34  & 31.77 & 19.06 & 32.26 &19.04 &26.41&30.04 & 31.83 & 32.15 & 30.46 & 28.78 & 30.58 & 32.06 & 30.63 & 31.72 & 33.71 & 29.91 & 36.36  & \textcolor{red}{41.25} & 40.14 & 22.55 & 22.44 & 30.97 & 30.56   & 41.37 \\
\hline
L.C.RL                & 22.64  & 41.40 & 56.19 & 40.36 & 42.47&37.81&50.87 & 38.96 & 37.75 & 56.99 & 37.77 & 59.75 & 40.25 & 55.26 & \textcolor{red}{61.05} & 37.75 & 47.81 & 56.16  & 49.58 & 54.62 & 37.68 & 13.88 & 16.89 & 46.74   & 47.71 \\
L.S.RL                & 46.37  & 8.59  & 13.49 & 32.97 &0.66 &39.72&36.75 & 39.94 & 44.46 & 38.99 & 45.60 & 39.46 & 37.46 & 43.10 & 43.32 & 46.24 & 54.90 & 50.65  & 54.87 & \textcolor{red}{55.19} & 0.00  & 31.56 & 32.43 & 37.88   & 37.04 \\
\rowcolor[HTML]{EFEFEF} 

MEAN                  & 29.82  & 31.47 & 43.27 & 38.12 &29.83 &38.39&46.60 & 39.26 & 39.78 & 51.54 & 40.14 & 53.61 & 39.40 & 51.58 & \textcolor{red}{55.69} & 40.32 & 49.95 & 54.49  & 51.18 & 54.79 & 26.29 & 19.23 & 21.59 & 44.06   & 44.48 \\
\hline
VRU.CI                & 61.39  & 44.64 & 57.83 & 46.37 &62.40 &55.51&57.36 & 62.53 & 61.07 & 59.96 & 53.76 & 59.66 & 45.84 & 62.38 & 62.38 & 63.31 & \textcolor{red}{68.54} & 64.72  & 54.96 & 59.38 & 51.25 & 58.65 & 58.41 & 63.30   & 63.43 \\
VRU.C                 & \textcolor{red}{88.15}  & 35.94 & 31.16 & 37.55 &80.92&37.23& 78.86 & 45.49 & 82.52 & 74.96 & 70.98 & 77.77 & 37.52 & 66.20 & 55.49 & 65.05 & 75.38 & 72.39  & 51.07 & 54.46 & 66.05 & 44.29 & 47.45 & 83.97   & 59.40 \\
V.CI                  & 62.59  & 36.25 & 20.80 & 33.05 & 83.10&21.84&52.24 & 24.14 & 39.31 & 36.17 & 25.52 & 42.53 & 36.51 & 27.64 & 28.47 & 26.93 & 34.48 & 26.17  & 23.54 & 28.28 & \textcolor{red}{86.70} & 29.31 & 18.72 & 44.60   & 26.09 \\
L.S.P                 & 58.69  & 24.44 & 44.81 & 24.06 &50.10 &49.75&52.56 & 53.25 & 58.75 & 58.13 & 60.75 & 57.50 & 24.56 & 45.00 & 54.00 & 57.25 & 62.75 & 58.75  & 58.75 & \textcolor{red}{71.00} & 32.00 & 54.50 & 47.06 & 52.25   & 60.63 \\
\rowcolor[HTML]{EFEFEF} 

MEAN                  & 67.47  & 34.80 & 38.98 & 34.76 & \textcolor{red}{68.23}&41.56&60.08 & 46.79 & 60.62 & 57.60 & 53.42 & 59.52 & 35.57 & 50.30 & 50.43 & 53.57 & 60.66 & 55.95  & 47.79 & 54.26 & 57.64 & 47.16 & 43.27 & 60.89   & 52.97 \\
\hline
VRU.B.                & 27.88  & 10.50 & 43.64 & 23.84 &29.00&43.43& 41.21 & 45.86 & 44.04 & 39.80 & 44.04 & 42.42 & 11.50 & 45.25 & 46.87 & 47.88 & 44.24 & 53.94  & \textcolor{red}{58.79} & 52.93 & 0.00  & 36.77 & 40.20 & 42.02   & 41.21 \\
V.B.                  & 34.00  & 10.50 & 21.50 & 10.75 &27.00&45.00& 33.00 & \textcolor{red}{48.00} & 39.00 & 44.00 & 40.00 & 44.00 & 11.50 & 37.00 & 41.25 & 47.00 & 47.00 & 46.50  & 45.00 & 47.00 & 0.25  & 45.75 & 40.00 & 44.00   & 45.00 \\
\rowcolor[HTML]{EFEFEF} 

MEAN                  & 30.61  & 10.50 & 33.74 & 17.99 &28.60&44.13& 37.54 & 46.82 & 41.79 & 41.68 & 42.23 & 43.13 & 11.50 & 41.56 & 44.36 & 47.49 & 45.47 & 50.61  & \textcolor{red}{52.63} & 50.28 & 0.11  & 40.78 & 40.11 & 42.91   & 42.91 \\
\hline
K.O.D                 & 26.62  & 54.52 & 44.42 & 53.93 &\textcolor{red}{76.01}&54.66 &59.49 & 69.73 & 62.41 & 59.49 & 58.17 & 69.73 & 55.50 & 71.33 & 69.87 & 73.53 & 72.94 & 63.44  & 70.60 & 75.72 & 74.11 & 66.95 & 64.75 & 75.72   & 74.55 \\
R.P.                  & 73.09  & \textcolor{red}{75.06} & 11.94 & 69.47 &62.16&31.18& 72.61 & 59.29 & 41.91 & 30.88 & 55.40 & 33.04 & 71.75 & 48.90 & 42.17 & 71.97 & 44.92 & 57.10  & 66.57 & 54.60 & 68.01 & 40.88 & 24.78 & 33.60   & 79.15 \\
D.E.                  & 48.58  & 38.88 & 54.30 & 40.75 &44.95& 46.47&48.78 & 54.85 & 51.68 & 61.29 & 61.19 & 55.92 & 40.86 & 48.38 & 50.83 & 46.36 & 63.04 & 60.33  & 53.61 & \textcolor{red}{63.70} & 47.68 & 53.47 & 59.27 & 46.60   & 58.15 \\
ST.R                  & 54.92  & 33.98 & 36.44 & 38.94 &64.08&44.49& 52.36 & 53.06 & 57.59 & 32.81 & 62.72 & 53.99 & 34.42 & 32.27 & 50.95 & 56.30 & \textcolor{red}{71.56} & 62.18  & 62.71 & 65.71 & 28.45 & 48.73 & 48.82 & 49.26   & 59.20 \\
\rowcolor[HTML]{EFEFEF} 

MEAN                  & 45.91  & 52.60 & 36.89 & 52.03 &64.24&46.47& 58.85 & 61.91 & 54.93 & 47.35 & 58.87 & 56.50 & 52.84 & 56.26 & 57.13 & 65.72 & 64.49 & 61.20  & 65.75 & \textcolor{red}{67.15} & 58.72 & 55.95 & 52.99 & 56.60   & 69.73 \\
\hline
LA.                   & \textcolor{red}{63.91}  & 22.55 & 24.17 & 23.49 &26.81&56.77& 40.26 & 41.45 & 30.21 & 32.43 & 33.62 & 41.11 & 24.43 & 46.81 & 35.15 & 62.55 & 55.40 & 56.09  & 43.83 & 44.85 & 11.74 & 39.74 & 50.98 & 53.36   & 59.83 \\
LO.                   & 36.24  & 9.31  & 20.20 & 6.73  &17.03&26.34& 24.95 & 39.41 & 46.93 & 46.34 & 39.80 & 41.39 & 7.33  & \textcolor{red}{68.91} & 46.14 & 48.51 & 50.10 & 57.23  & 57.82 & 54.85 & 18.22 & 31.49 & 47.72 & 29.90   & 50.89 \\
\rowcolor[HTML]{EFEFEF} 

MEAN                  & 55.60  & 18.57 & 22.98 & 18.45 &23.87&47.62& 35.65 & 40.83 & 35.24 & 36.61 & 35.48 & 41.19 & 19.29 & 53.45 & 38.45 & \textcolor{red}{58.33} & 53.81 & 56.43  & 48.04 & 47.86 & 13.69 & 37.26 & 50.00 & 46.31   & 57.14 \\
\rowcolor[HTML]{EFEFEF} 
\hline
\hline

TOTAL                 & 30.93  & 35.17 & 35.86 & 38.67 &43.24&44.97& 45.45 & 48.58 & 49.40 & 49.97 & 50.75 & 51.73 & 38.76 & 50.21 & 51.62 & 53.20 & 56.51 & 54.23  & 56.00 & \textcolor{red}{58.00} & 33.87 & 44.90 & 47.45 & 49.83   & 57.39
 \\ \hline
\end{tabular}
}
\end{table*}

\section{Detailed Results}
The detailed results of the tertiary tasks are listed in \cref{detailed_results_each}, and more motion planning results are presented in \cref{more_tra}.
\section{Examples of \textbf{VLADBench}}

\begin{figure}[!htbp]
\vspace{-0.3cm}
  \centering
\includegraphics[width=\linewidth]{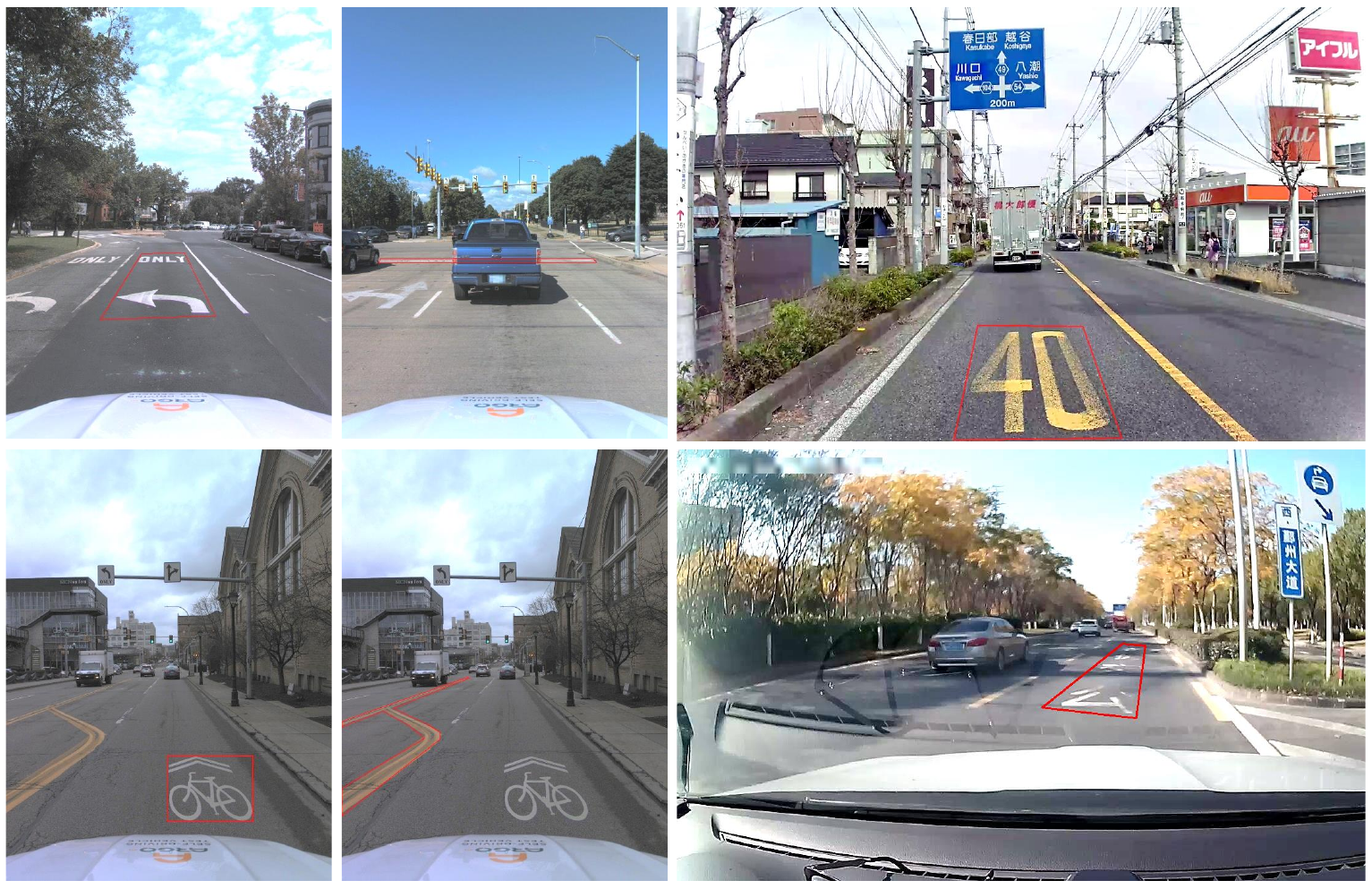}
   \vspace{-0.7cm}
   \caption{Examples of Pavement Marking.}
   \label{fig:sup_pm}
   \vspace{-0.3cm}
\end{figure}

\label{sec:sup_tasks}
\noindent \textbf{Pavement Marking} refers to markings painted or applied to roads, pavements, and other traffic areas to guide drivers, cyclists, and pedestrians. In this task, the questions include the type (a total of 20) and meaning of various pavement markings, including lane lines, arrows, symbols, text, etc., as illustrated by the samples in \cref{fig:sup_pm}.

\begin{figure}[!htbp]
\vspace{-0.3cm}
\centering
\includegraphics[width=\linewidth]{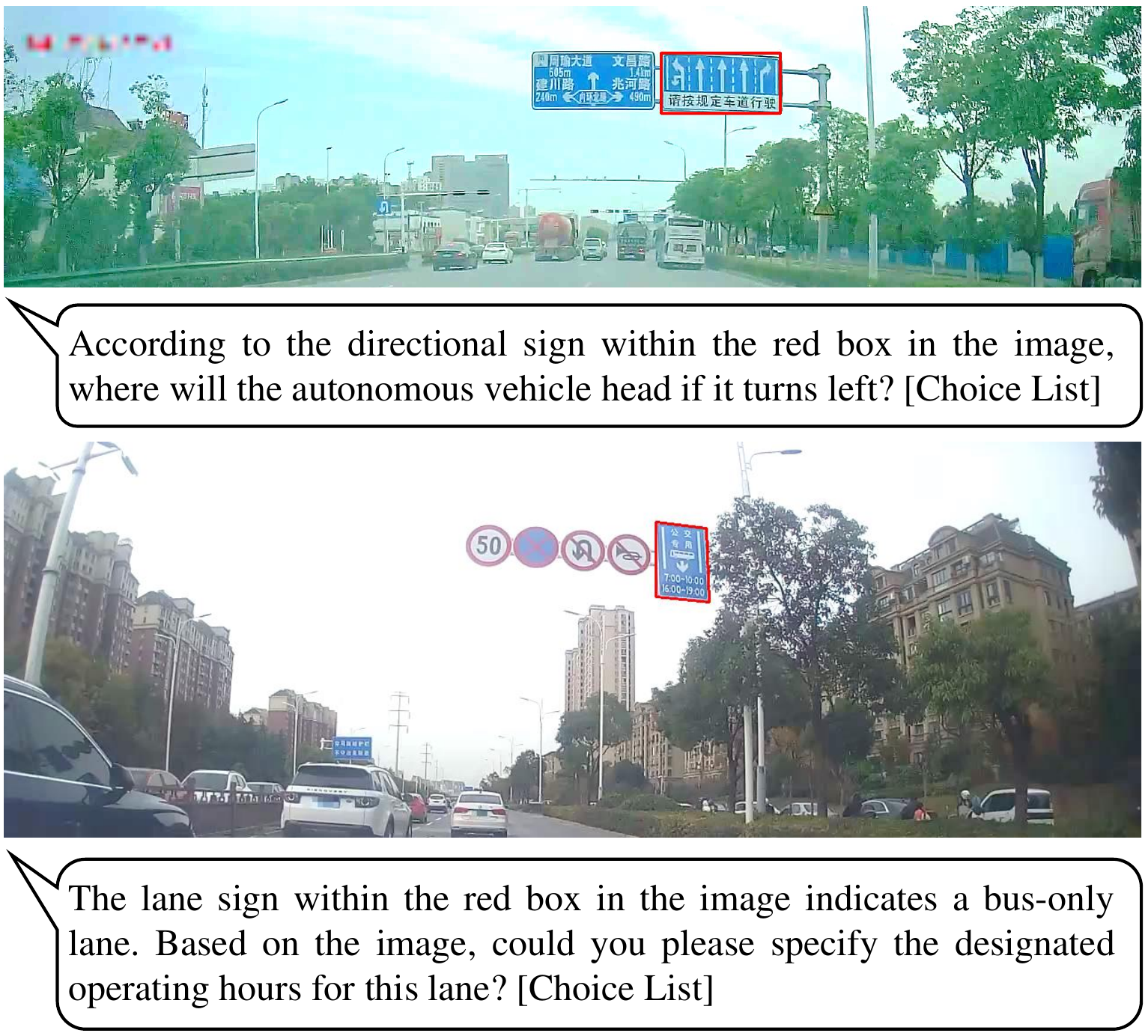}
   \vspace{-0.7cm}
   \caption{Examples of Traffic Sign.}
   \label{fig:sup_ts}
   \vspace{-0.3cm}
\end{figure}

\noindent \textbf{Traffic Sign} refers to a visual display placed at the roadside or above a road to inform drivers, cyclists, and pedestrians about the road, its conditions, and traffic regulations. The category includes lane signs, directional signs, regulatory signs, prohibitory signs, warning signs and construction signs, encompassing a total of 168 types. Specifically, we present two important and intriguing questions in \cref{fig:sup_ts}, which are designed to conduct an in-depth evaluation of VLM-based AD.

\begin{figure}[!htbp]
  \centering
\includegraphics[width=\linewidth]{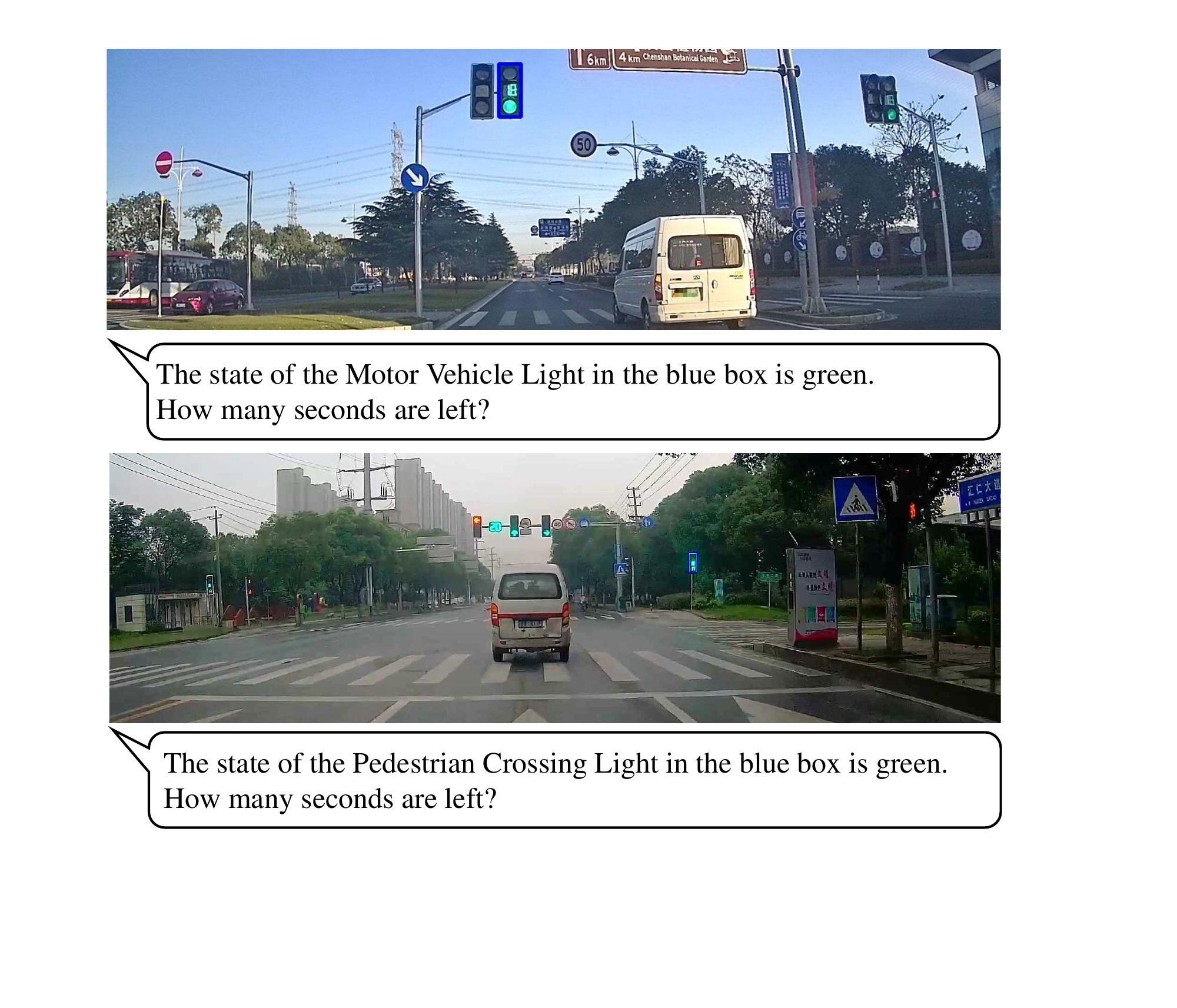}
   \vspace{-0.70cm}
   \caption{Examples of Traffic Light.}
   \label{fig:sup_tl}
   \vspace{-0.3cm}
\end{figure}

\noindent \textbf{Traffic Light} refers to a signaling device used to control the flow of traffic at intersections and other locations. The types include the motor vehicle light, non-motorized light, pedestrian crossing light, lane light, and arrow light. The traffic light statuses include red, yellow, green and malfunction. In addition to the type and status, we also incorporate questions regarding countdown timers, as illustrated in \cref{fig:sup_tl}.

\begin{figure}[!htbp]
  \centering
\includegraphics[width=\linewidth]{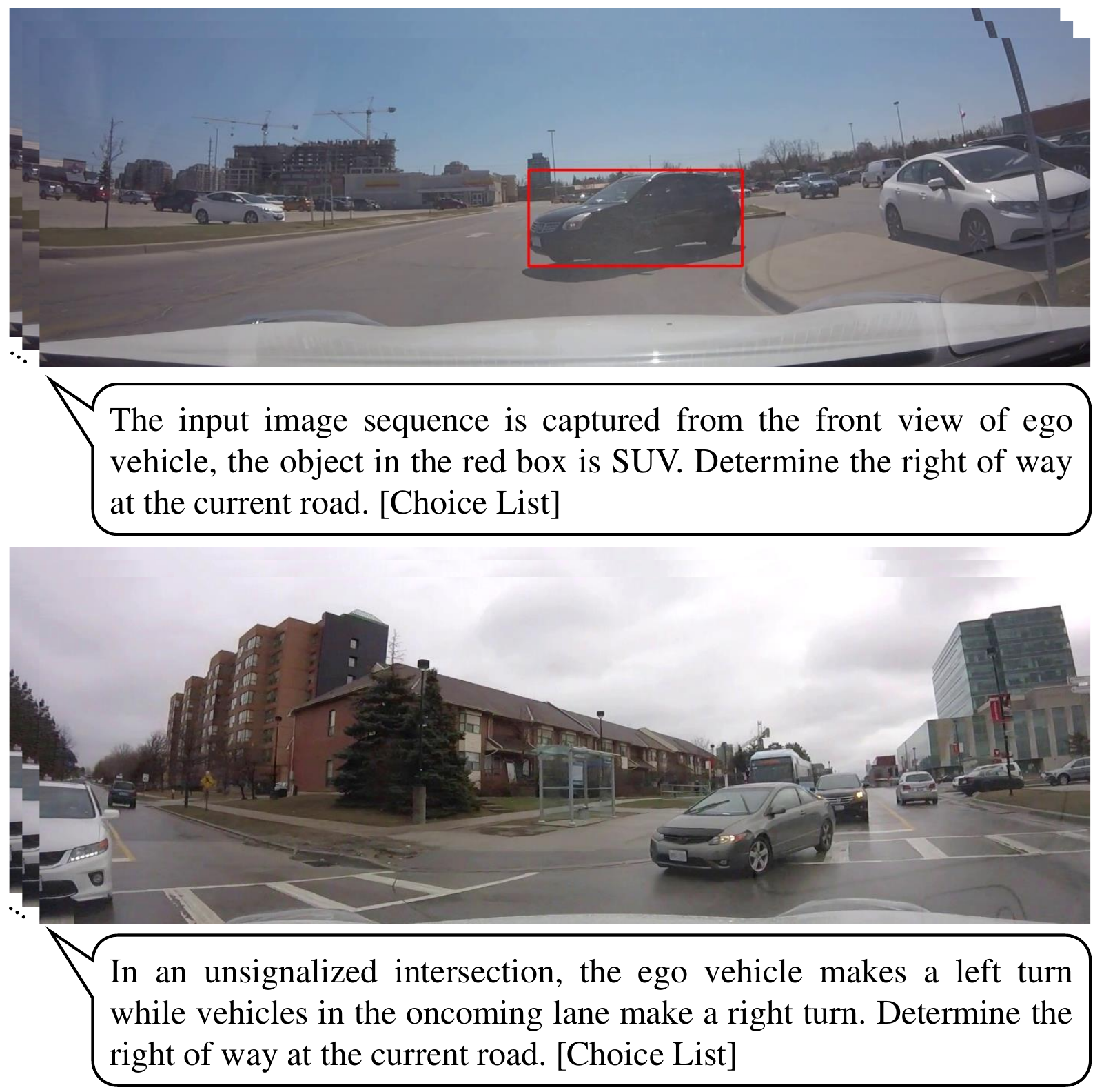}
   \vspace{-0.70cm}
   \caption{Examples of Right Of Way.}
   \label{fig:sup_row}
   \vspace{-0.3cm}
\end{figure}

\noindent \textbf{Right Of Way} refers to the legal right of a person or vehicle to proceed before an ego vehicle at an intersection or other point on a road. Examples are illustrated in \cref{fig:sup_row}. Additionally, we provide descriptions of the actions to take when there are no visual prompts.

\begin{figure}[!ht]
  \centering
\includegraphics[width=\linewidth]{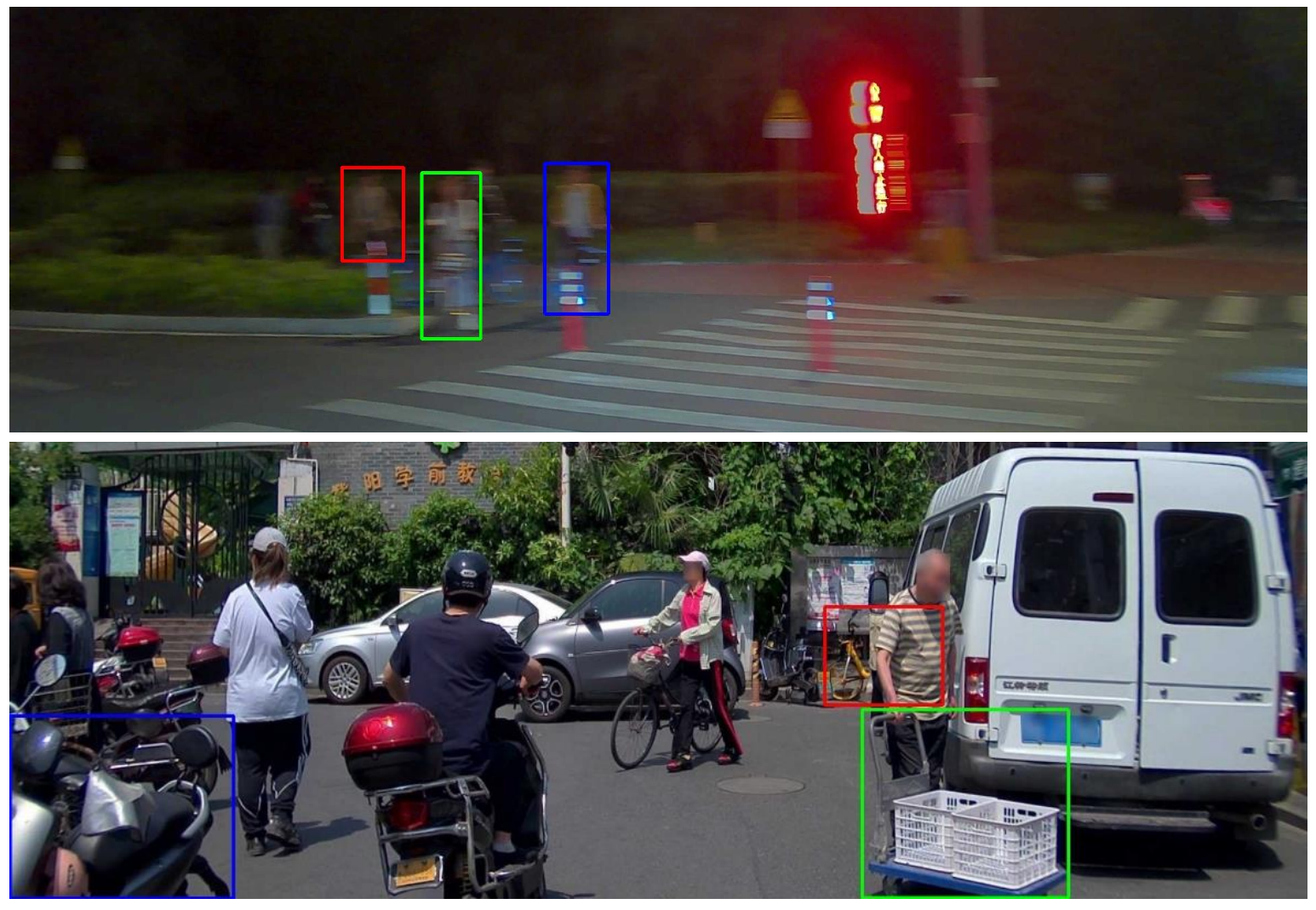}
   \vspace{-0.70cm}
   \caption{Examples of VRU Recognition.}
   \label{fig:sup_vru_r}
   \vspace{-0.3cm}
\end{figure}
\noindent \textbf{VRU Recognition} refers to the classification and detection for vulnerable road users (VRUs). There are 8 types of VRUs: moped, tricycle, cart, cyclist, bicycle, stroller, motorcycle, and wheelchair. Note that these VRUs are selected from corner cases. Examples are shown in \cref{fig:sup_vru_r}.

\begin{figure}[!ht]
  \centering
\includegraphics[width=\linewidth]{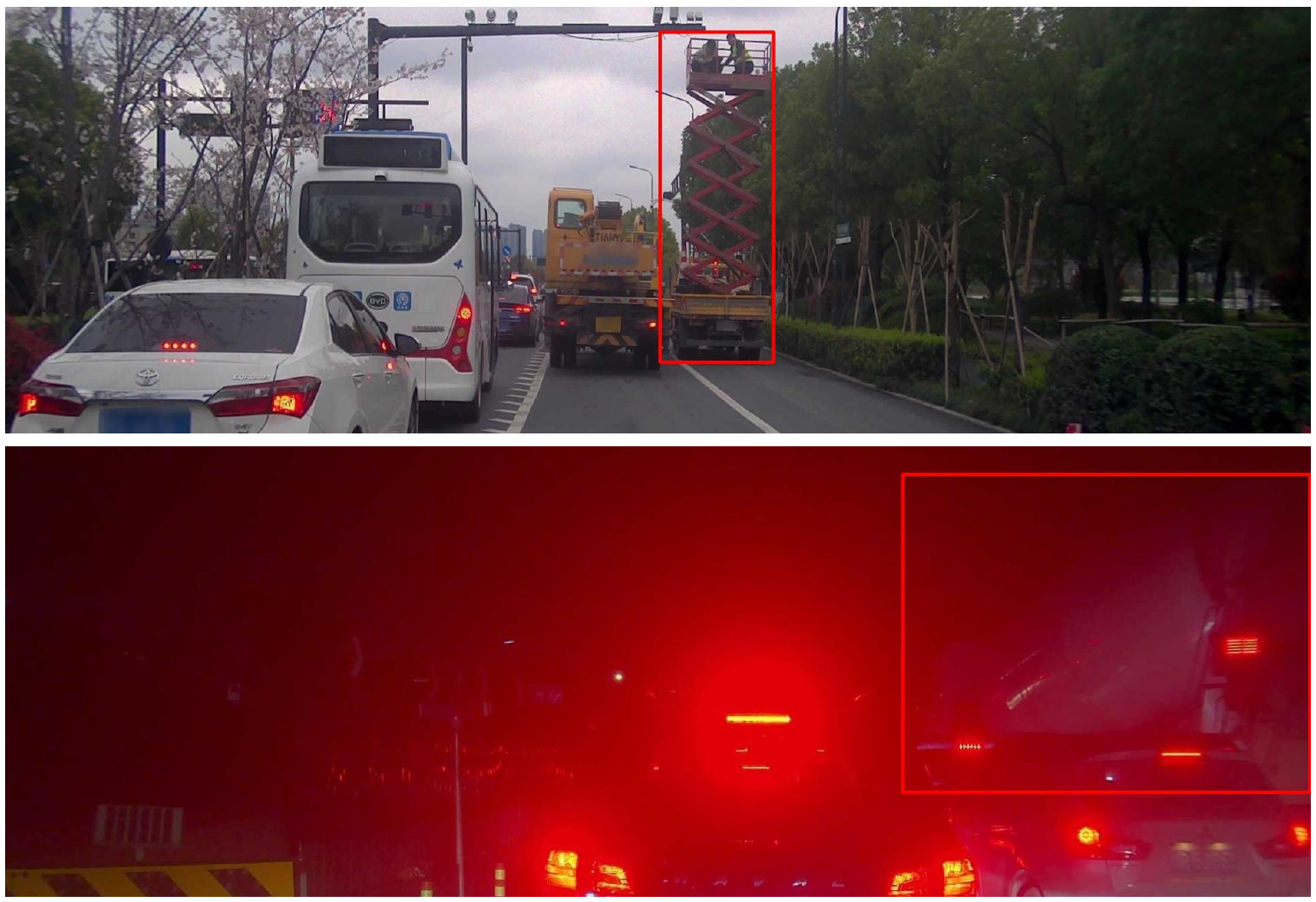}
   \vspace{-0.70cm}
   \caption{Examples of Vehicle Recognition.}
   \label{fig:sup_v_r}
   \vspace{-0.3cm}
\end{figure}

\noindent \textbf{Vehicle Recognition} refers to the classification and detection for 5 types of corner-case level vehicles, including car, construction vehicle, truck, bus, and sanitation vehicle. Examples are shown in \cref{fig:sup_v_r}.

\begin{figure}[!htbp]
  \centering
\includegraphics[width=\linewidth]{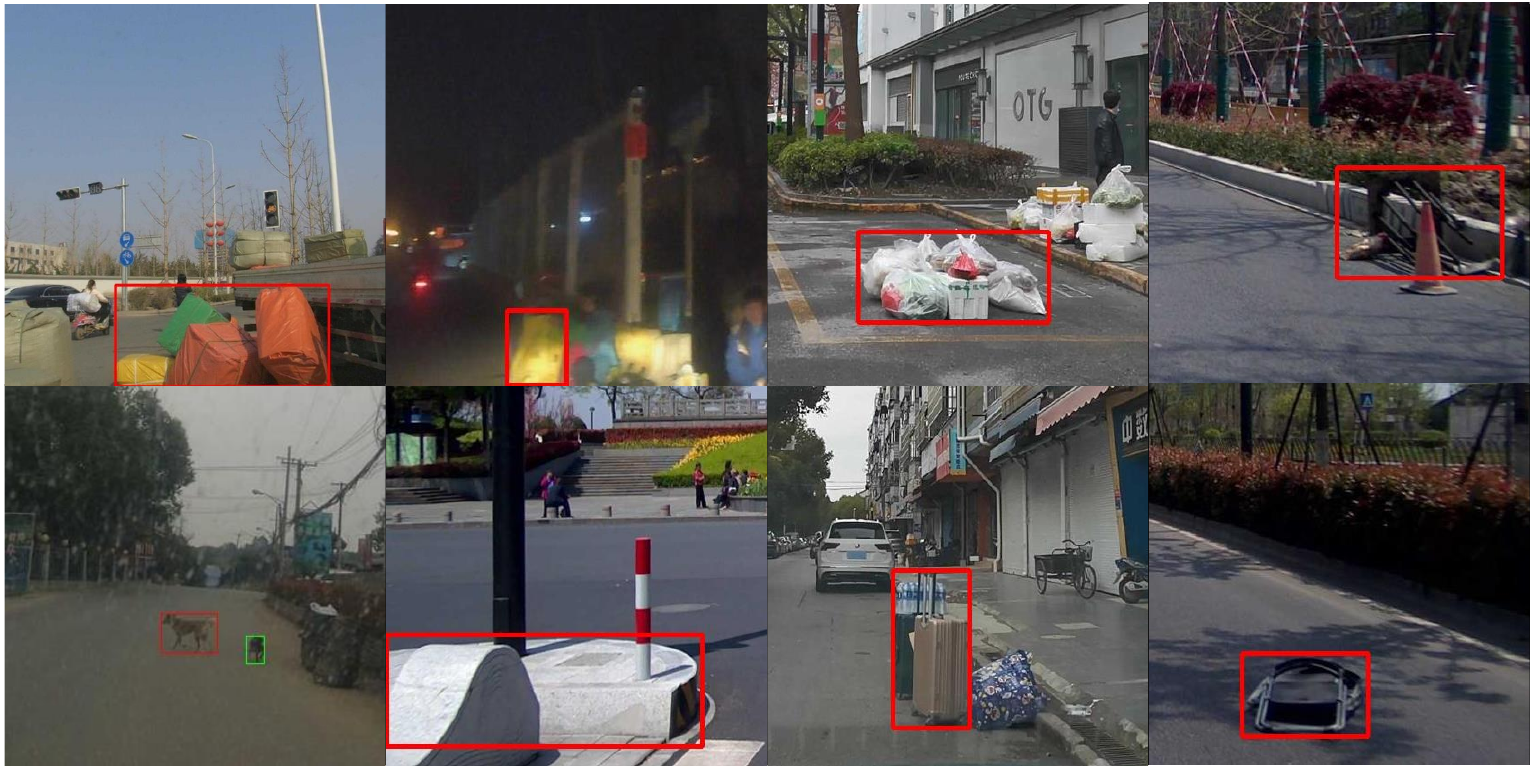}
   \vspace{-0.70cm}
   \caption{Examples of Obstruction Recognition.}
   \label{fig:sup_os}
   \vspace{-0.3cm}
\end{figure}
\noindent \textbf{Obstruction Recognition} refers to the classification and detection for the 19 types of corner-case level obstacles, including debris, suitcase, concrete block, plastic bag, chair, machinery, phone booth, dustbin, basket, stone, garbage, tire, carton, cardboard, garbage bag, traffic cone, traffic box, traffic island, and dog. 
These obstacles, often found unexpectedly on the road or in parking areas, pose significant risks to the safety and efficiency of autonomous vehicles. The ability to accurately detect and classify such obstructions is critical for ensuring smooth navigation and avoiding accidents in dynamic environments. Examples are shown in \cref{fig:sup_os}.

\begin{figure}[!htbp]
  \centering
\includegraphics[width=\linewidth]{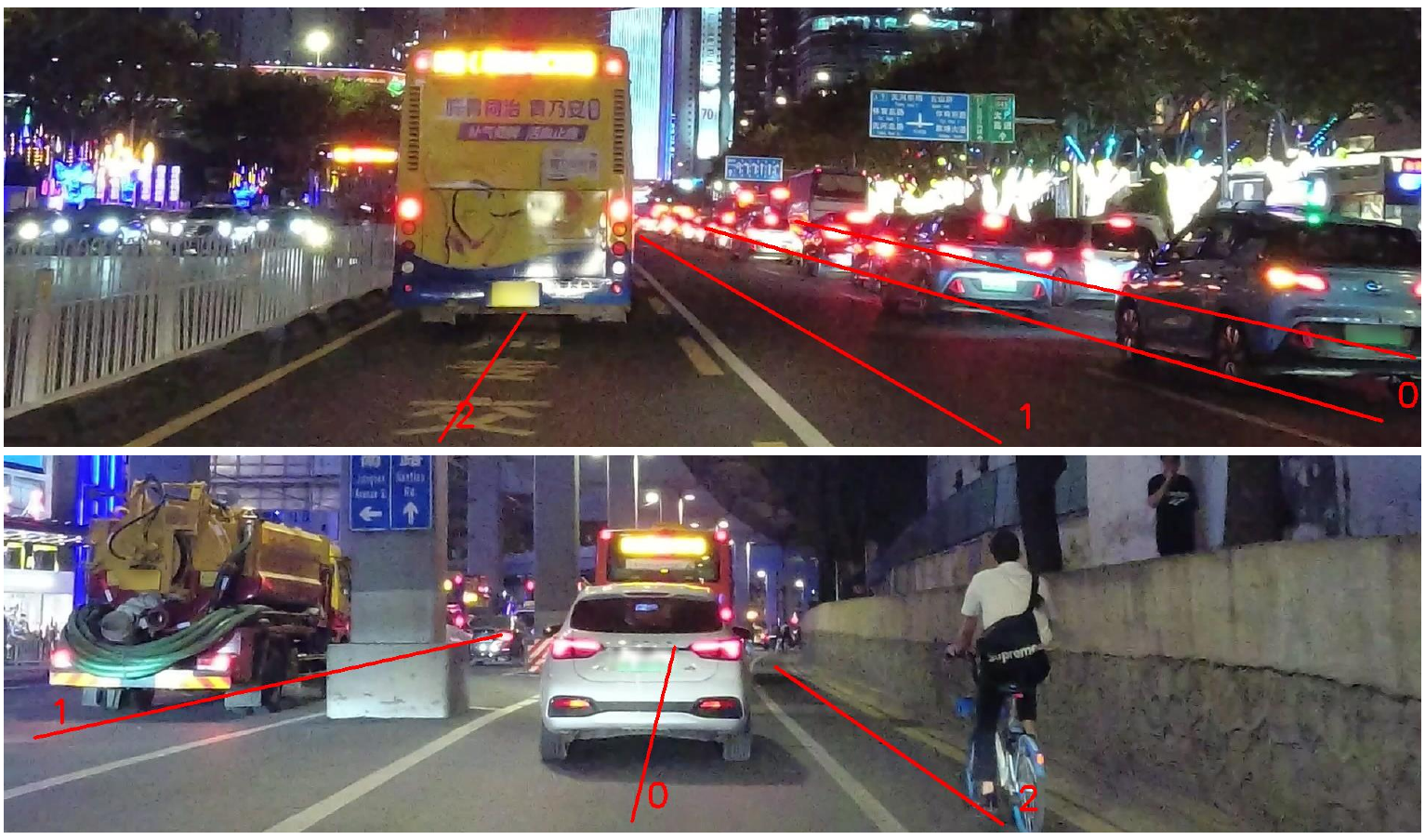}
   \vspace{-0.70cm}
   \caption{Examples of Lane Recognition.}
   \label{fig:sup_lr}
   \vspace{-0.3cm}
\end{figure}

\noindent \textbf{Lane Recognition} involves identifying various types of lanes, such as motorized vehicle lanes, non-motorized vehicle lanes, emergency lanes, dedicated bus lanes, ETC-exclusive lanes, and sidewalks. Additionally, we increase the complexity of the lane recognition task by distinguishing between opposing lanes and classifying motor vehicle lanes as fast, slow, or regular. The added complexity helps validate the ability to navigate complex roadways, enhancing its decision-making process in various traffic conditions. Accurate lane recognition is crucial for safe lane changes, merging, and ensuring the vehicle stays within the correct lane. Examples are shown in \cref{fig:sup_lr}.


\begin{figure}[!htbp]
  \centering
\includegraphics[width=\linewidth]{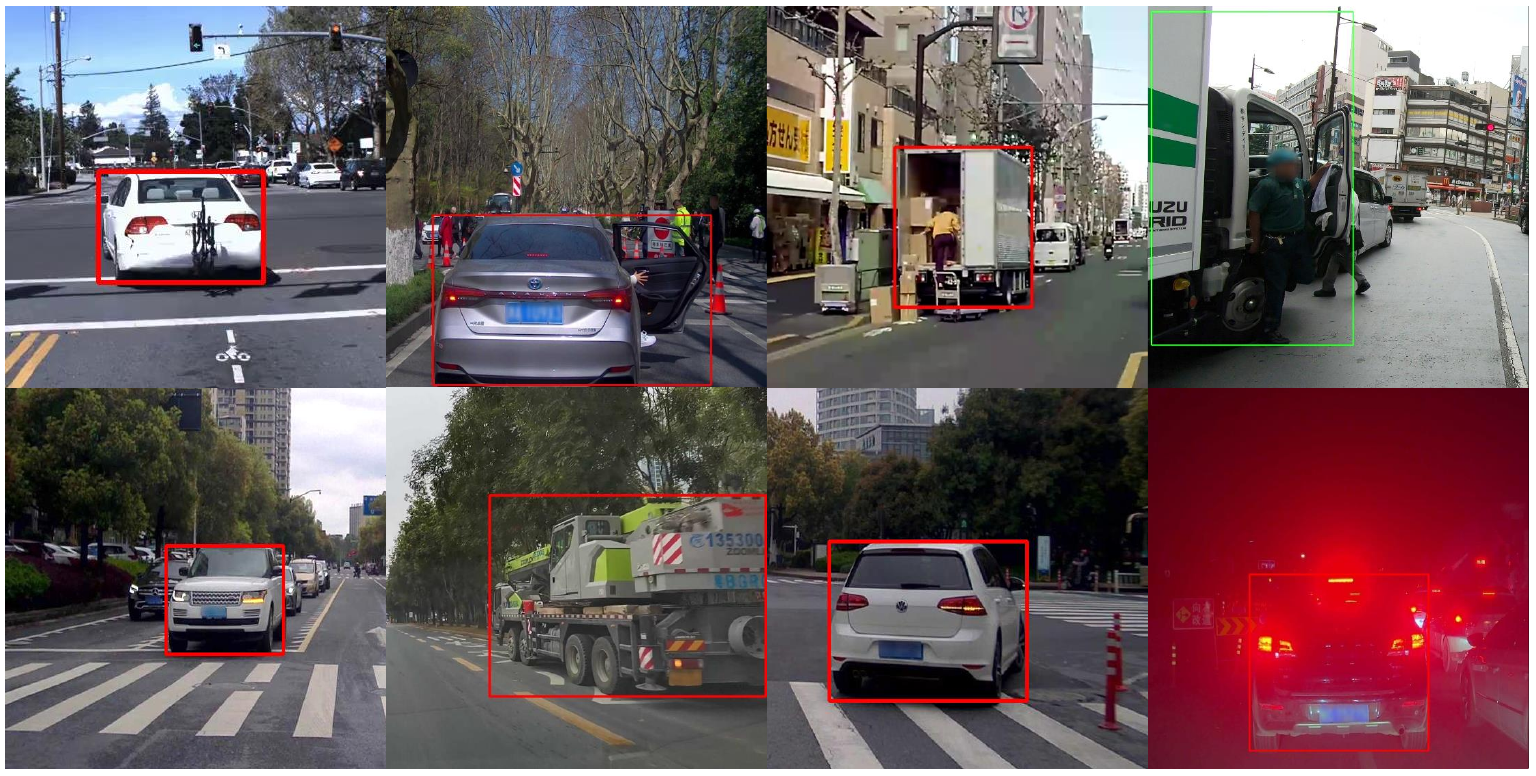}
   \vspace{-0.70cm}
   \caption{Examples of Vehicle Status.}
   \label{fig:sup_vs}
   \vspace{-0.3cm}
\end{figure}

\noindent \textbf{Vehicle Status} focuses on the light state and operational status, referring to observable exterior conditions of a target vehicle. There are 17 possible statuses, including right turn light on, left turn light on, brake light on, hazard lights on, empty car light on, passenger light on, right door open, left door open, trunk open, trunk open for loading, right door open for boarding, left door open for boarding, rear compartment door open for loading, all compartment doors open, construction work, accident scene, and cargo hanging from trunk. The exterior status provides critical information about the current state of the target vehicle, which is essential for autonomous systems to make accurate decisions in real-time traffic scenarios. For instance, recognizing whether doors are open or if hazard lights are on helps the system assess the vehicle’s intent or possible hazards. Examples are shown in \cref{fig:sup_vs}.

\begin{figure}[!htbp]
  \centering
\includegraphics[width=\linewidth]{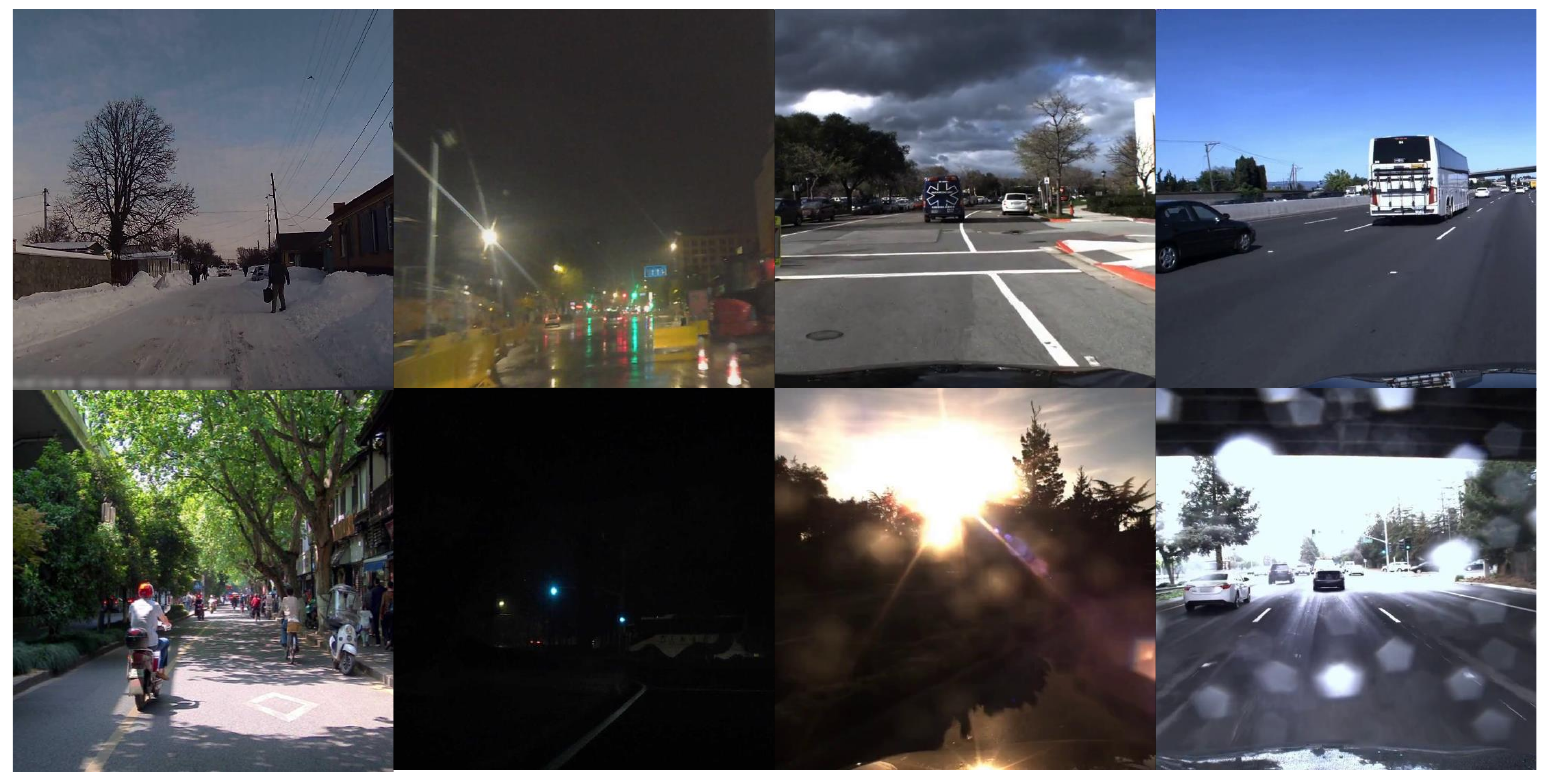}
   \vspace{-0.70cm}
   \caption{Examples of Weather \& Light .}
   \label{fig:sup_wl}
   \vspace{-0.3cm}
\end{figure}
\noindent \textbf{Weather \& Light} refer to the driving conditions in the scenarios. 
The questions are first addressed in terms of daytime, nighttime, and dawn\&dusk conditions, followed by considerations of light and weather. The light conditions comprise diffuse, backlit, snowy, shadowed light, bright, low, very low, and dark. The weather conditions consist of overcast, clear, rainy, cloudy, and snowy. These environmental factors significantly impact the perception and decision-making processes of AD systems, as they affect visibility, road conditions, and overall driving safety. For instance, driving in low-light or snowy conditions requires the vehicle to adjust its speed and navigation strategy. Examples are shown in \cref{fig:sup_wl}.

\begin{figure}[!htbp]
  \centering
\includegraphics[width=\linewidth]{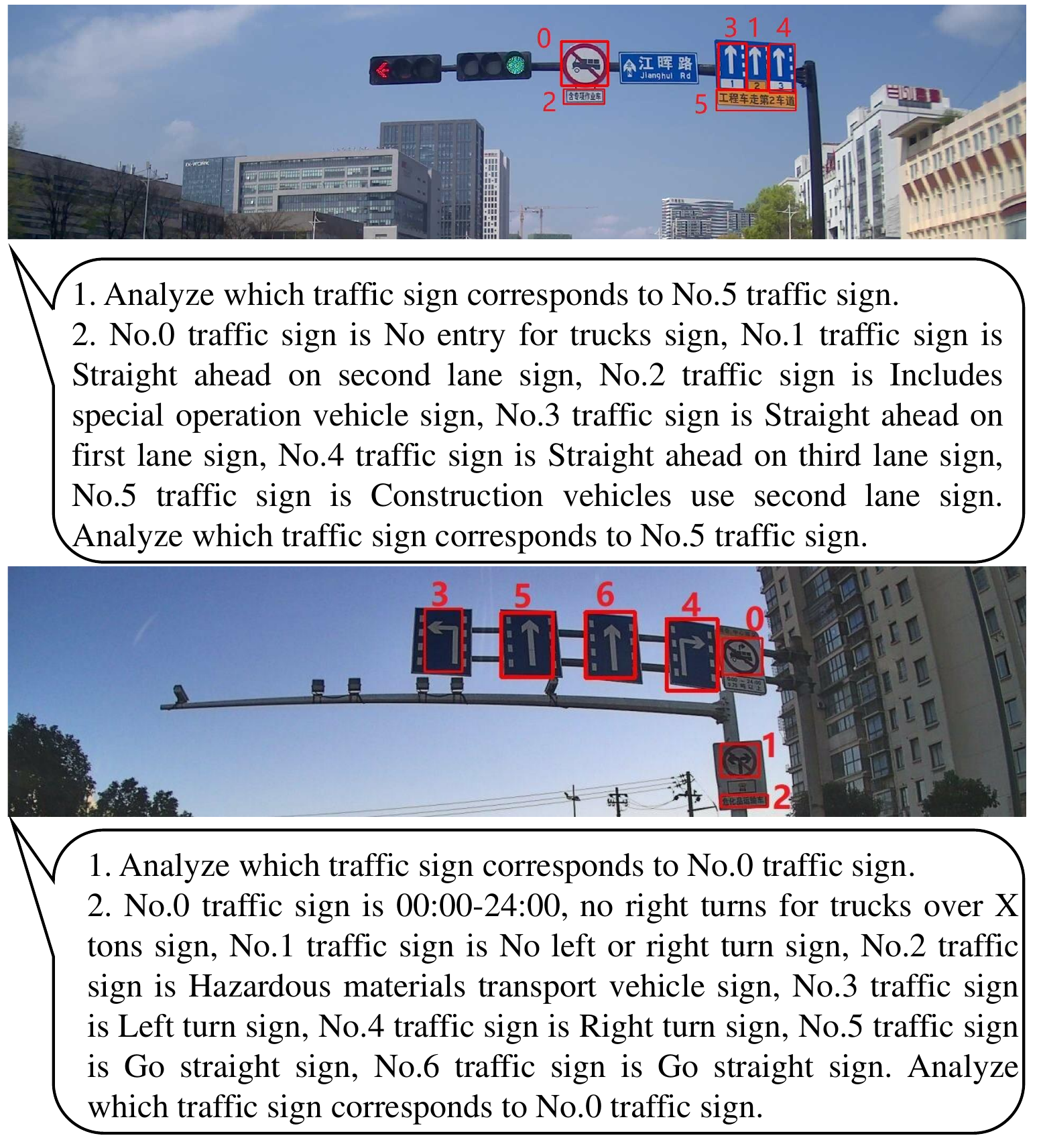}
   \vspace{-0.70cm}
   \caption{Examples of Sign-Sign Relation.}
   \label{fig:sup_ss}
   \vspace{-0.3cm}
\end{figure}

\noindent \textbf{Sign-Sign Relation} refers to the traffic graph connections between different signs. In the image, traffic signs are highlighted with red boxes, and the closest red numbers indicate the corresponding traffic sign identifiers. The Sign-Sign Relation task analyzes which sign corresponds to a specified sub-sign, capturing the hierarchical or contextual relationships between signs. Examples are shown in \cref{fig:sup_ss}.
\begin{figure}[!htbp]
  \centering
\includegraphics[width=\linewidth]{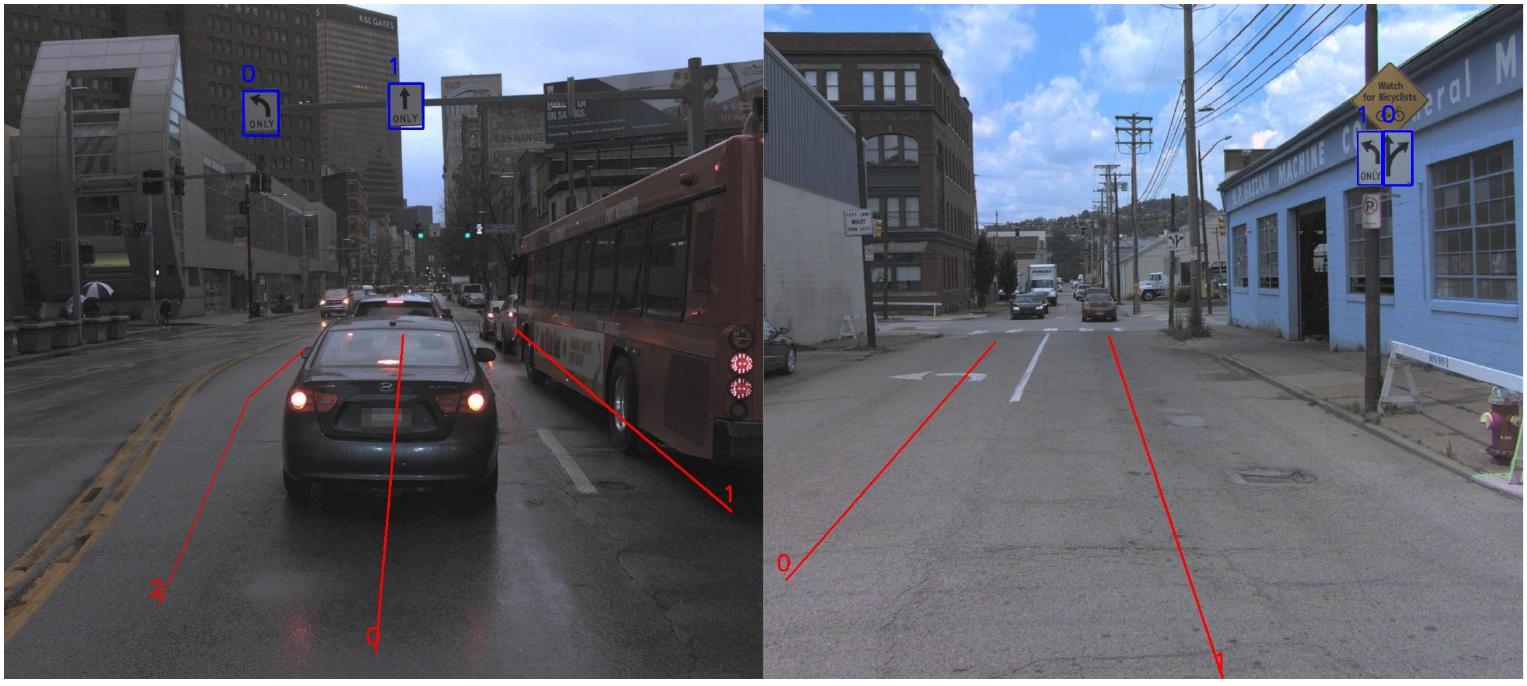}
   \vspace{-0.70cm}
   \caption{Examples of Lane-Sign Relation.}
   \label{fig:sup_lanes}
   \vspace{-0.3cm}
\end{figure}

\noindent \textbf{Lane-Sign Relation} refers to the traffic graph about the lanes and traffic signs. In the image, the lanes are represented by red lines, with the adjacent red numbers indicating the lane identification. Traffic signs are depicted within blue boxes, with the nearest blue numbers indicating traffic sign numbers. The Lane-Sign Relation task focuses on determining which lane corresponds to a specified traffic sign, enabling the system to interpret lane-specific rules or guidance effectively. Examples are shown in \cref{fig:sup_lanes}.

\begin{figure}[!htbp]
  \centering
\includegraphics[width=\linewidth]{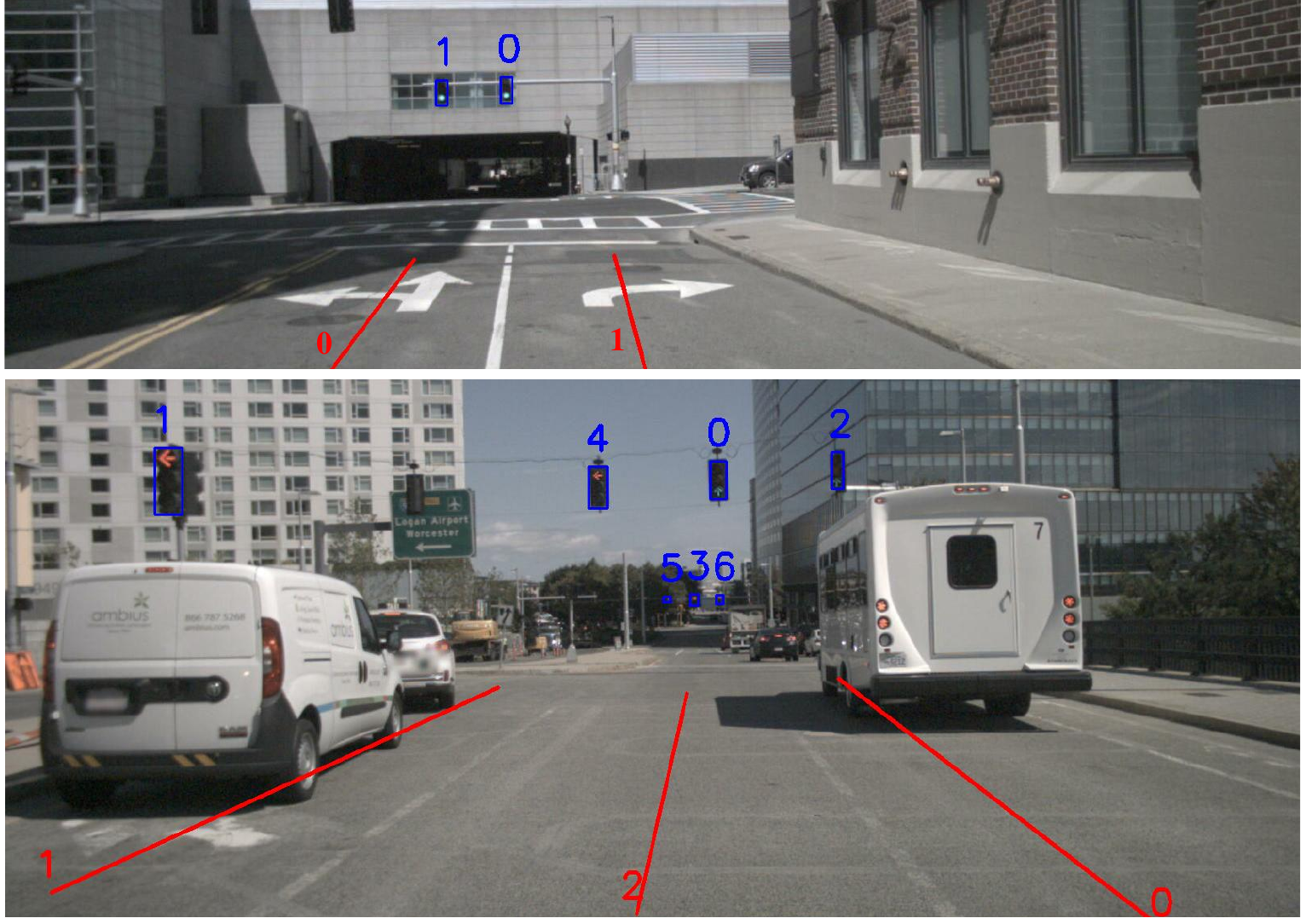}
   \vspace{-0.70cm}
   \caption{Examples of Light-Lane Relation.}
   \label{fig:sup_lanel}
   \vspace{-0.3cm}
\end{figure}
\noindent \textbf{Light-Lane Relation} refers to the traffic graph about the lanes and lights. In the image, the lanes are represented by red lines, with the adjacent red numbers indicating the lane identification. Traffic lights are represented by blue boxes, with the adjacent blue numbers indicating the traffic light identification. The Light-Lane Relation task analyzes which lane corresponds to a specified traffic light, facilitating an understanding of how traffic signals regulate specific lanes. Examples are shown in \cref{fig:sup_lanel}.

\begin{figure}[!htbp]
  \centering
\includegraphics[width=\linewidth]{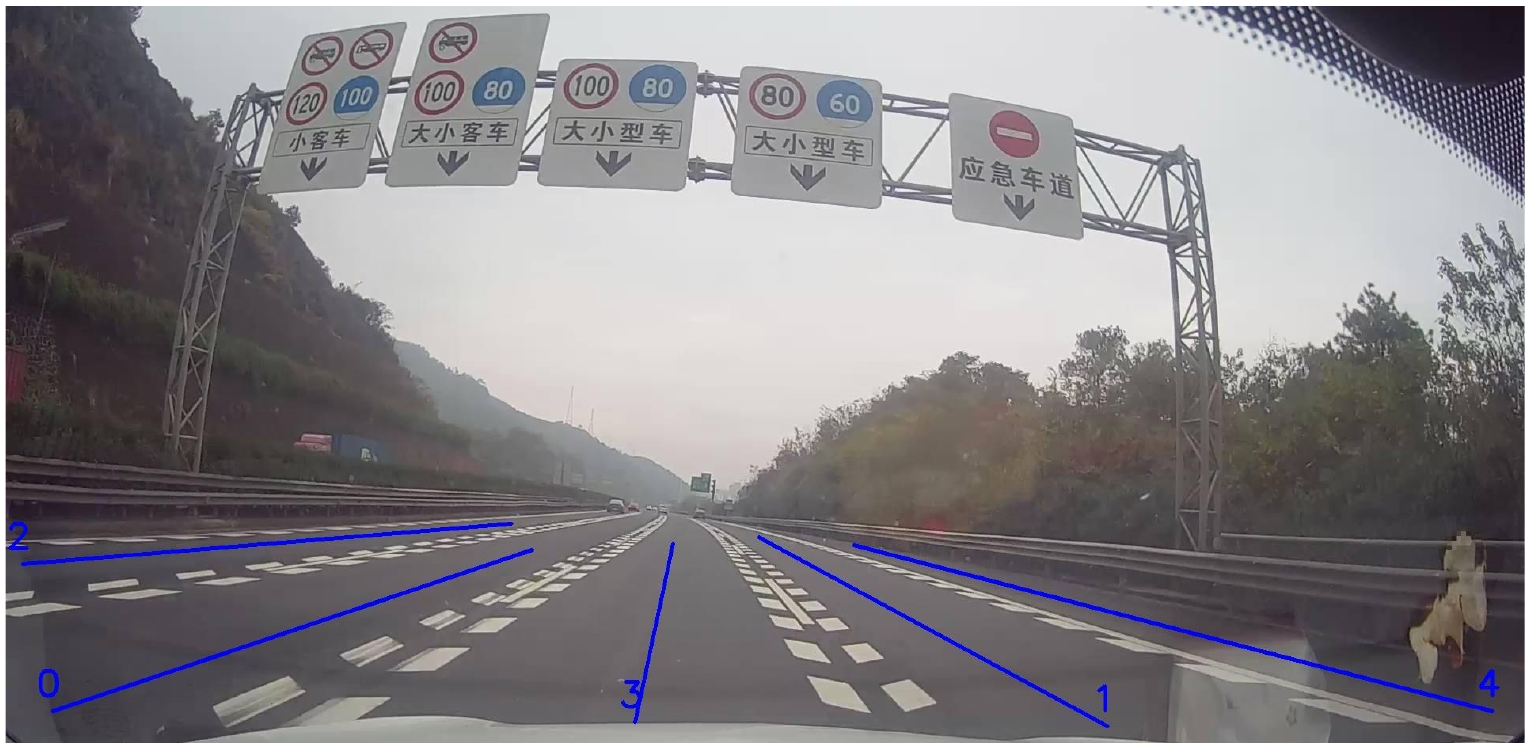}
   \vspace{-0.70cm}
   \caption{Examples of Lane Speed Relation.}
   \label{fig:sup_ls}
   \vspace{-0.3cm}
\end{figure}

\noindent \textbf{Lane Speed Relation} involves analyzing the low-speed and high-speed limits for a specified lane, based on the traffic signs present in the image. This task requires identifying the speed-related traffic signs and associating them with the relevant lanes to determine the permissible speed range. Examples are shown in \cref{fig:sup_ls}.
\begin{figure}[!htbp]
  \centering
\includegraphics[width=\linewidth]{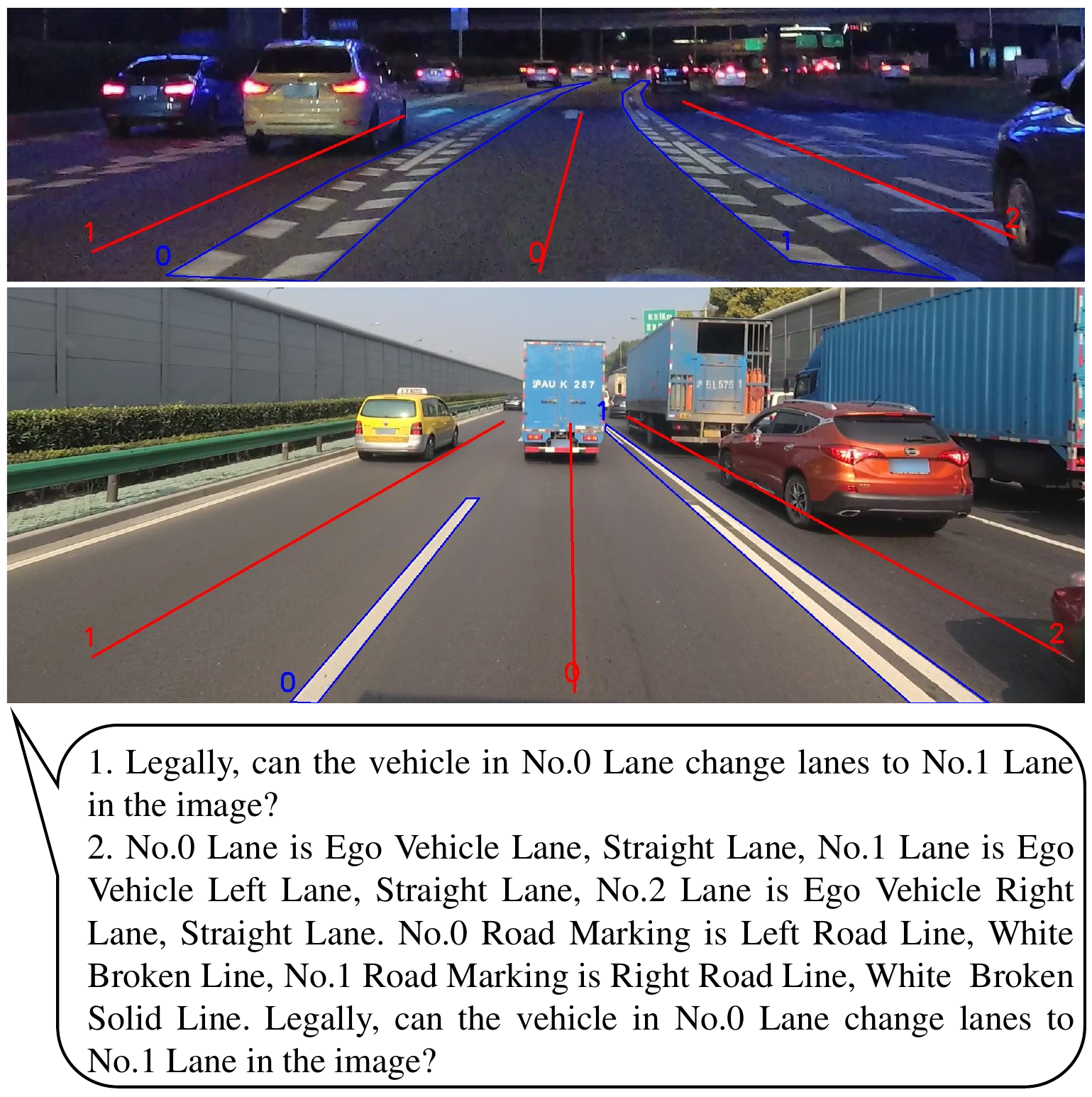}
   \vspace{-0.70cm}
   \caption{Examples of Lane Change Relation.}
   \label{fig:sup_lc}
   \vspace{-0.3cm}
\end{figure}

\noindent \textbf{Lane Change Relation} analyzes the permissibility and rules governing lane changes. This task is based on road markings, represented here by a blue box, and involves determining whether a lane change from one lane to another is allowed. Understanding lane-change relationships is critical for autonomous systems to safely navigate dynamic traffic environments, such as highways or multi-lane roads, where precise adherence to road markings is necessary to avoid collisions and ensure smooth traffic flow. Examples are shown in \cref{fig:sup_lc}.

\begin{figure}[!htbp]
  \centering
\includegraphics[width=\linewidth]{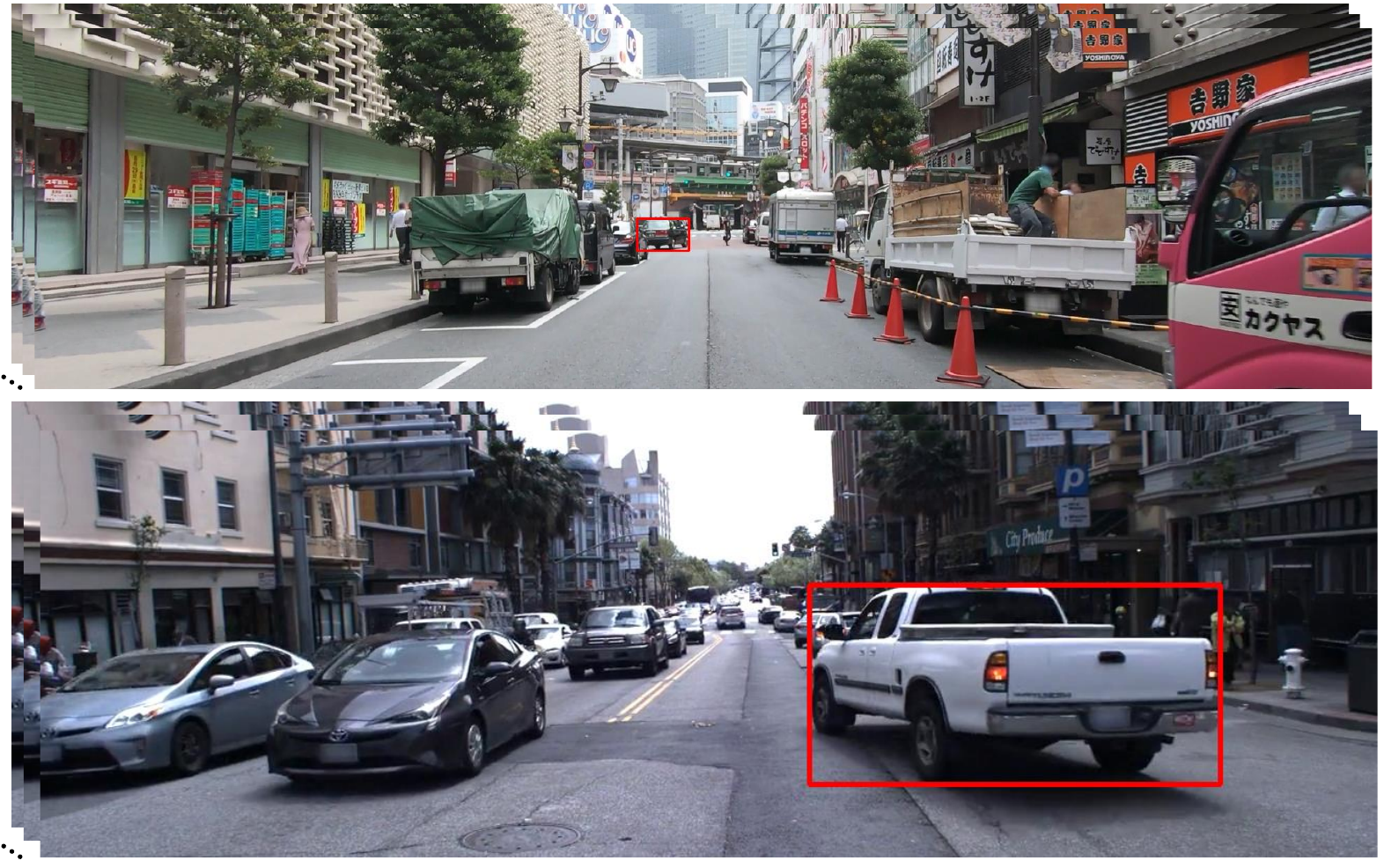}
   \vspace{-0.70cm}
   \caption{Examples of Vehicle Cut-in.}
   \label{fig:sup_vci}
   \vspace{-0.3cm}
\end{figure}
\noindent \textbf{Vehicle Cut-in} refers to the task of judging whether a target vehicle intends to merge from an adjacent lane or other areas into the lane of the ego vehicle, and analyzing the motivation behind the behavior. Examples are shown in \cref{fig:sup_vci}.

\begin{figure}[!htbp]
  \centering
\includegraphics[width=\linewidth]{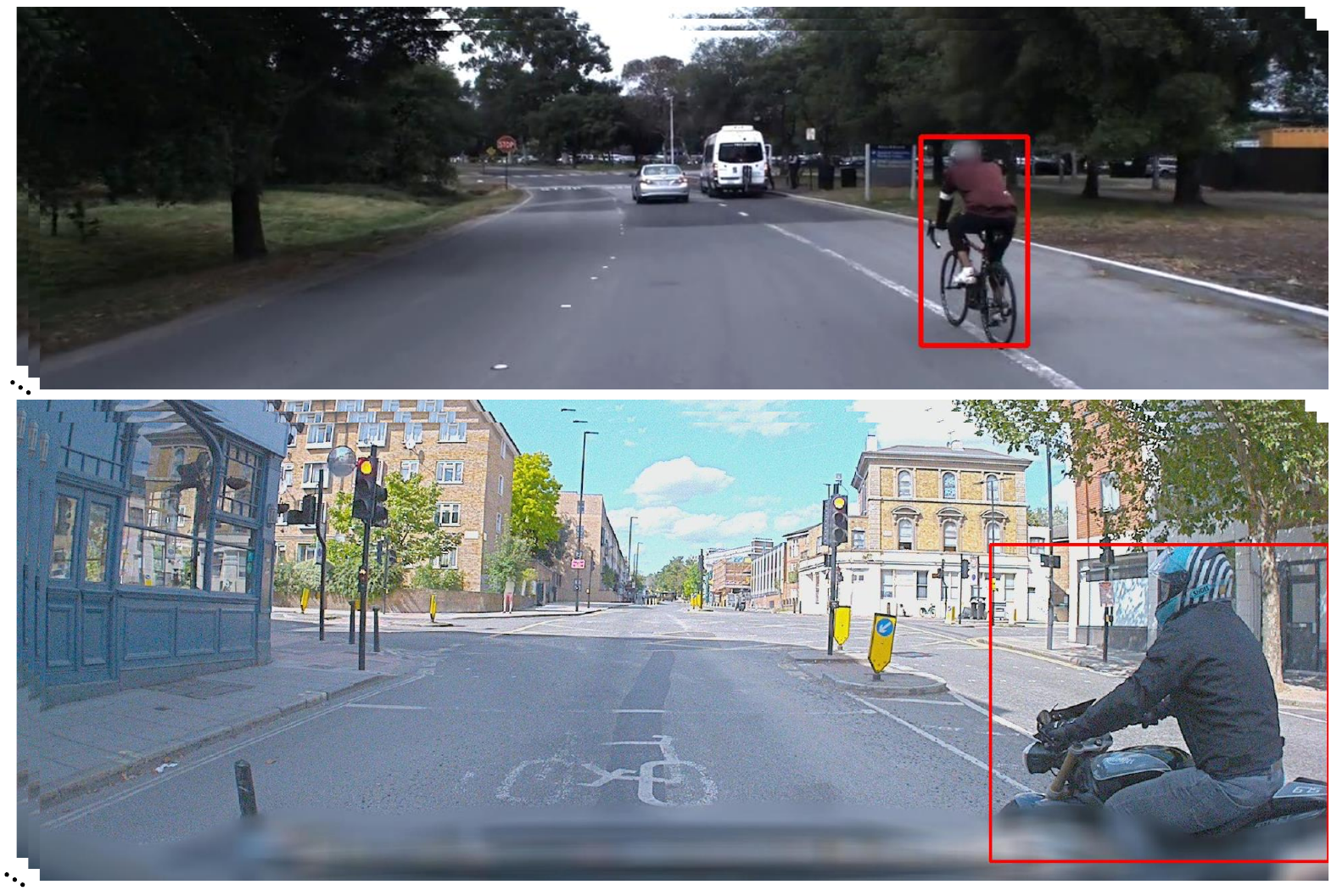}
   \vspace{-0.70cm}
   \caption{Examples of VRU Cut-in.}
   \label{fig:sup_VRUCI}
   \vspace{-0.3cm}
\end{figure}

\noindent \textbf{VRU Cut-in} refers to the task of judging whether a target VRU intends to merge from a different lane into the lane of the ego vehicle, and analyzing the motivation behind the behavior. Examples are shown in \cref{fig:sup_VRUCI}.

\begin{figure}[!htbp]
  \centering
\includegraphics[width=\linewidth]{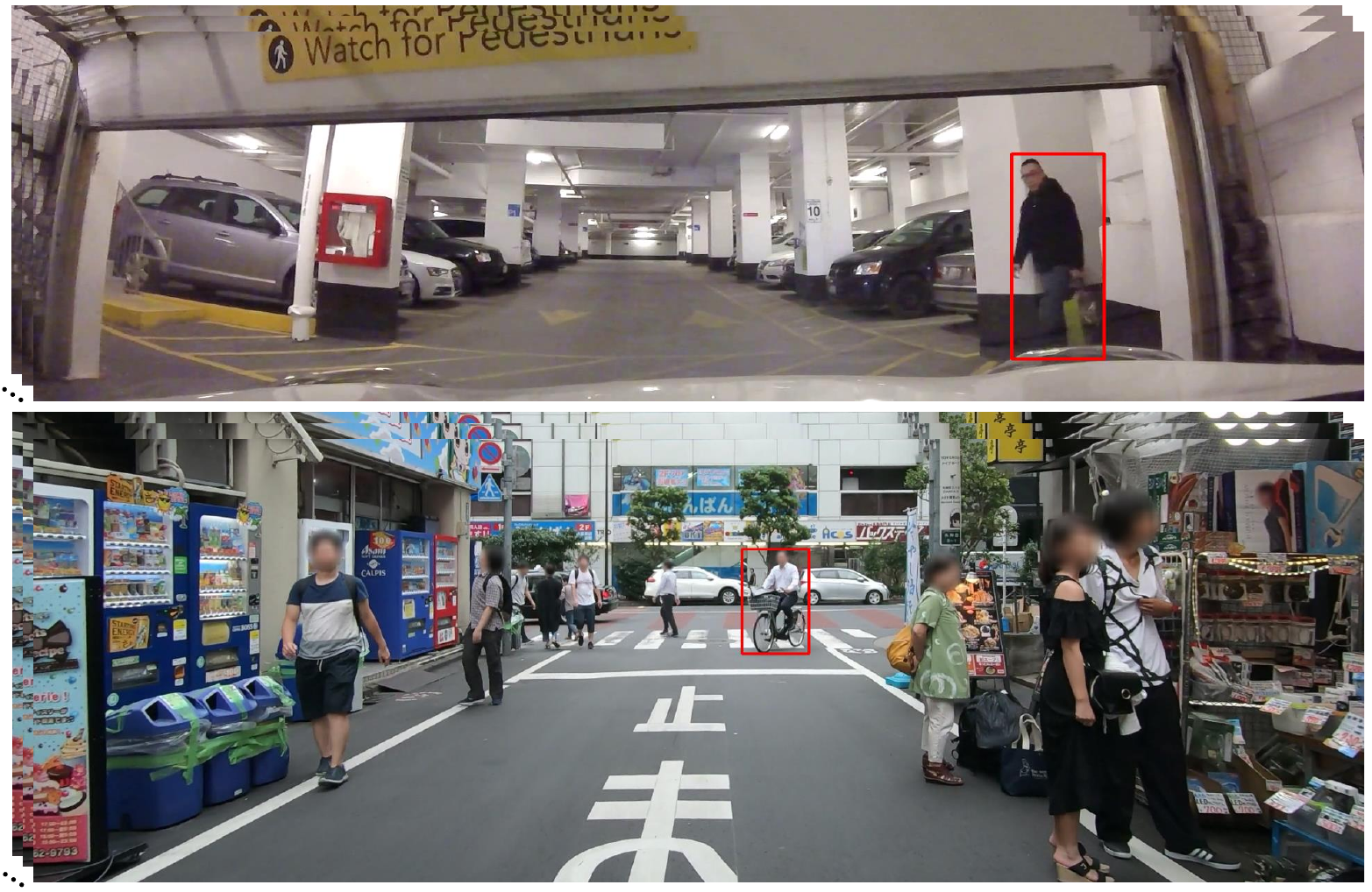}
   \vspace{-0.70cm}
   \caption{Examples of VRU Cross.}
   \label{fig:vru_c}
   \vspace{-0.3cm}
\end{figure}
\noindent \textbf{VRU Cross} refers to determining whether a VRU intends to cross laterally from one side to the other across the ego vehicle's path of travel, and analyzing the motivation behind the behavior. Examples are shown in \cref{fig:vru_c}.

\begin{figure}[!htbp]
  \centering
\includegraphics[width=\linewidth]{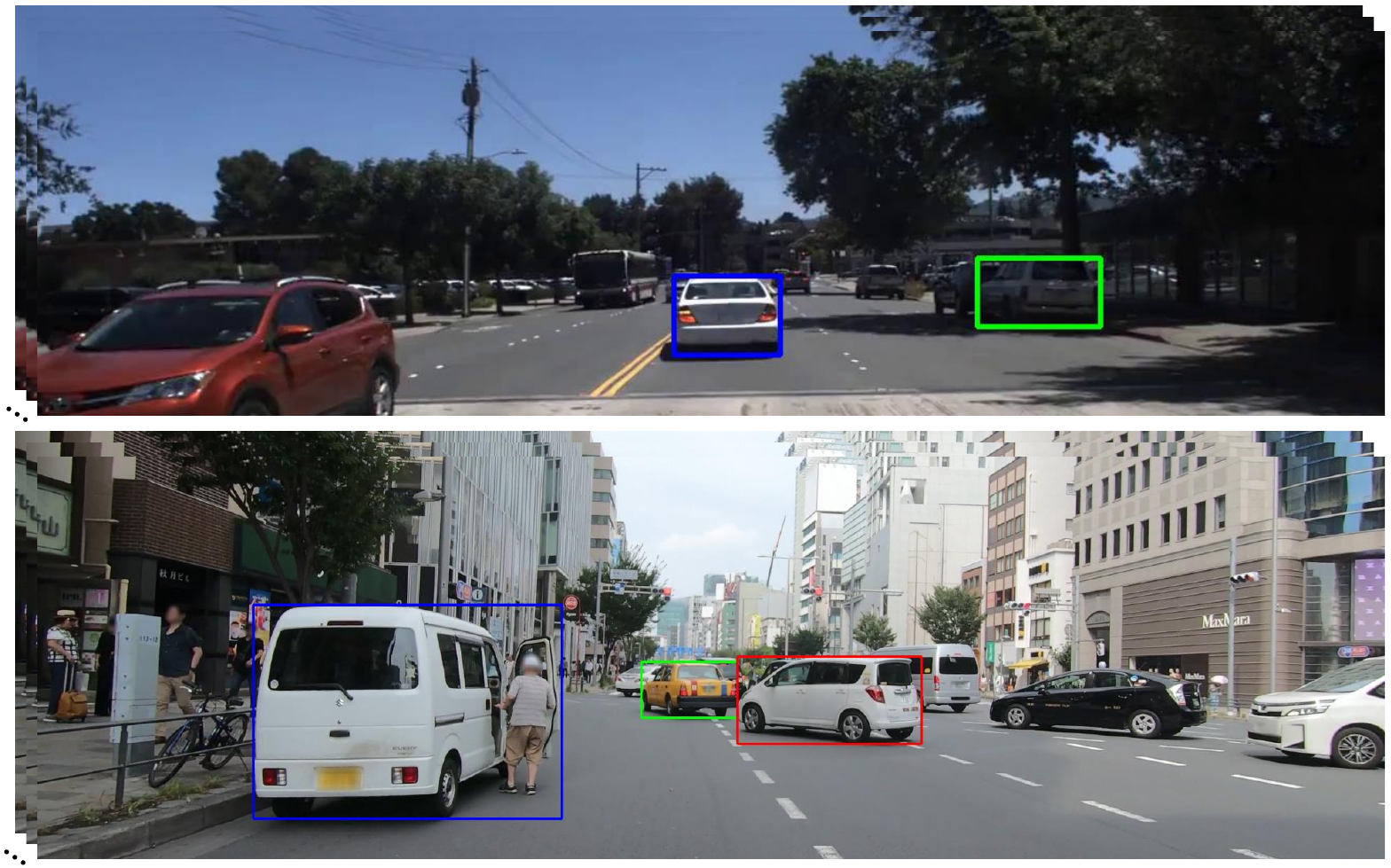}
   \vspace{-0.70cm}
   \caption{Examples of Long-Short Parking.}
   \label{fig:lsp}
   \vspace{-0.3cm}
\end{figure}
\noindent \textbf{Long-Short Parking} focuses on analyzing the parking time of the target vehicle. Whether the target vehicle is considered to be long-term or short-term parking (e.g., waiting for the traffic light, yielding, ever-changing passengers) is determined by whether the ego vehicle needs to perform a detour or execute an escape maneuver. Examples are shown in \cref{fig:lsp}.

\begin{figure}[!htbp]
  \centering
\includegraphics[width=\linewidth]{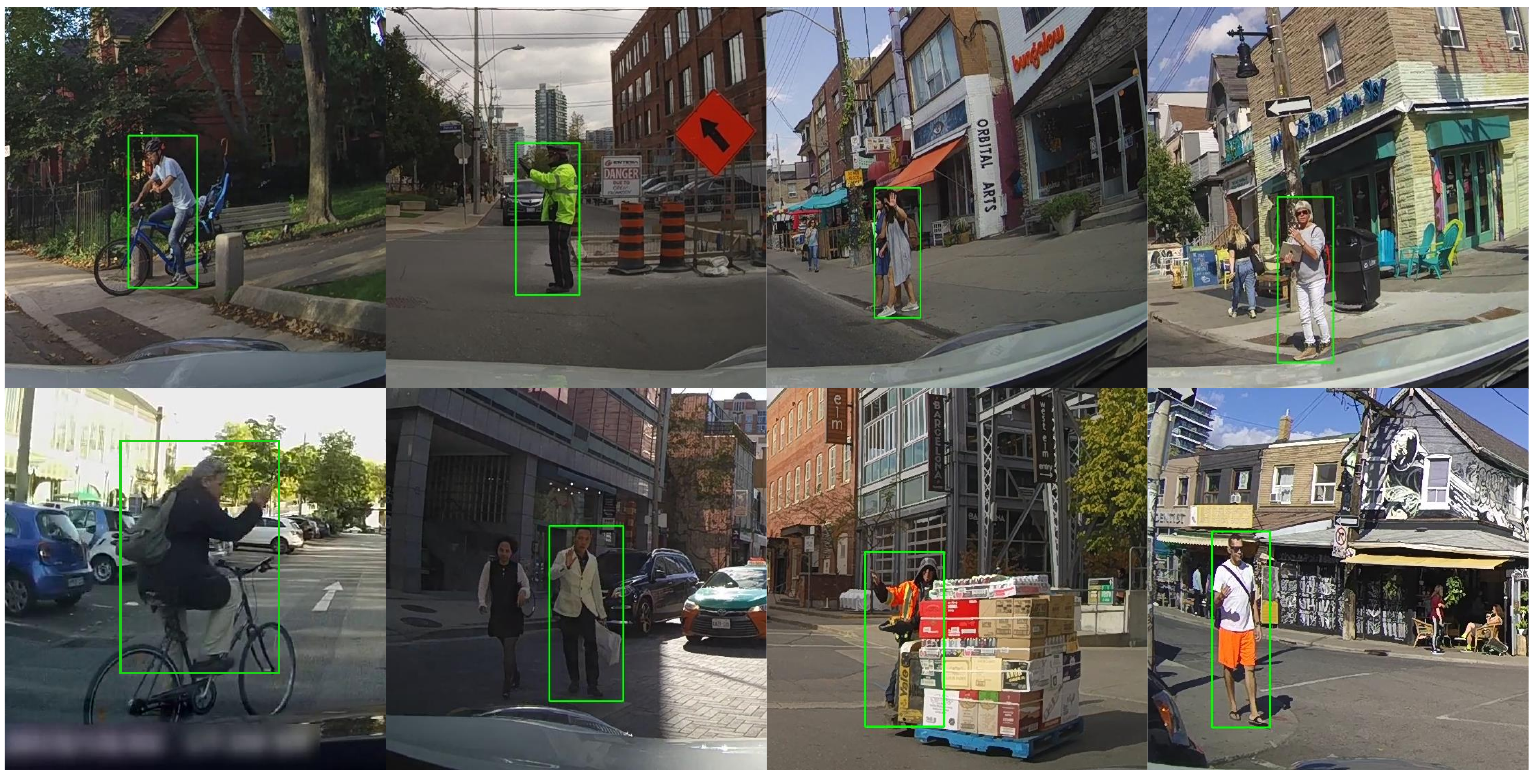}
   \vspace{-0.70cm}
   \caption{Examples of the pedestrian gesture of VRU Behavior.}
   \label{fig:vru_b}
   \vspace{-0.3cm}
\end{figure}
\noindent \textbf{Vehicle \& VRU Behavior} focuses on describing events that have occurred, with an emphasis on understanding and interpreting behaviors in the traffic environment. Vehicle behavior is characterized by longitudinal and lateral movements, capturing actions such as acceleration, braking, and lane changes. VRU behavior encompasses critical maneuvers, including cut-in and crossing actions, which are essential for predicting potential conflicts. Additionally, pedestrian gesture analysis is incorporated to assess claims of right of way, providing a deeper evaluation of interactions between pedestrians and vehicles. Examples are shown in \cref{fig:vru_b}.
\begin{figure}[!htbp]
  \centering
\includegraphics[width=\linewidth]{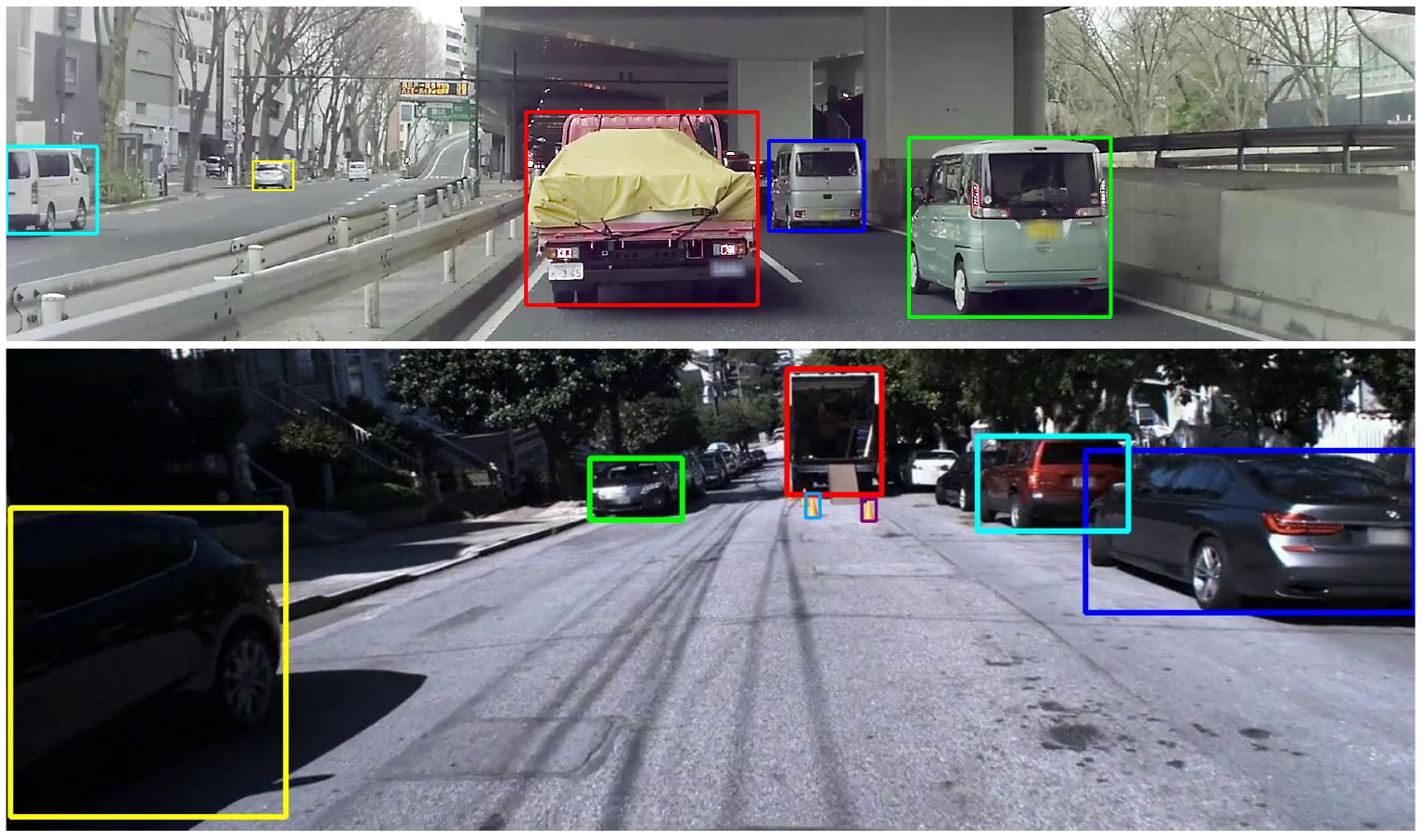}
   \vspace{-0.70cm}
   \caption{Examples of Key Object Detection.}
   \label{fig:kod}
   \vspace{-0.3cm}
\end{figure}

\noindent \textbf{Key Object Detection} refers to the identification of objects that play a critical role in determining the vehicle’s ability to maintain its current trajectory or safely execute lane changes to the left or right. These key objects may include vehicles, obstacles, or environmental elements that directly or indirectly influence driving decisions and safety. Examples are shown in \cref{fig:kod}.

\begin{figure}[!htbp]
  \centering
\includegraphics[width=\linewidth]{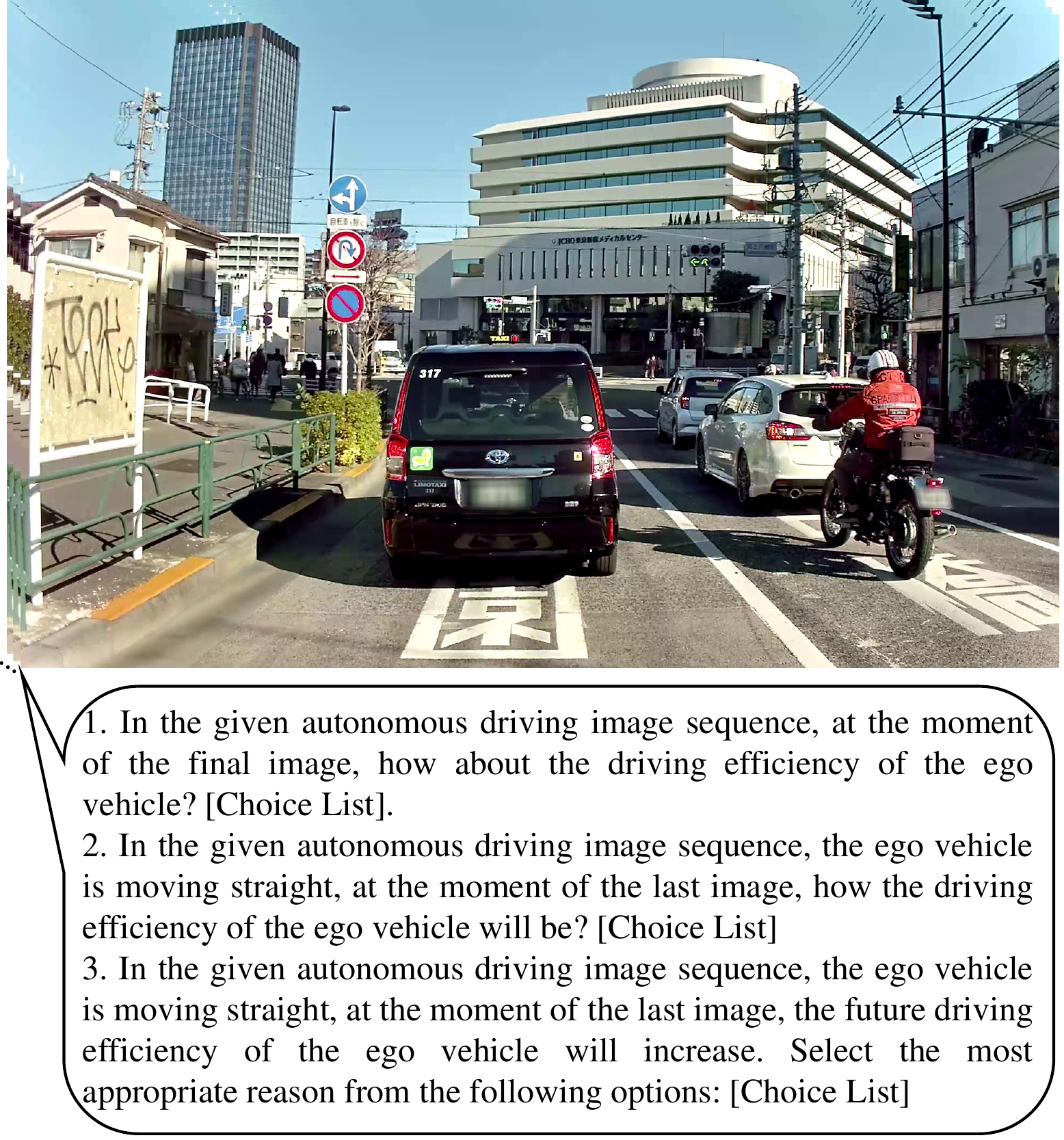}
   \vspace{-0.70cm}
   \caption{Examples of Drive Efficiency.}
   \label{fig:de}
   \vspace{-0.3cm}
\end{figure}

\noindent \textbf{Drive Efficiency} aims to evaluate the operational effectiveness of ego vehicles in relation to traffic congestion levels. The evaluation framework considers three aspects: the current driving efficiency under prevailing congestion conditions, projected efficiency changes as congestion evolves, and the factors influencing these changes. By emphasizing congestion as a key metric, this task provides insights into optimizing driving strategies in dense traffic environments. Examples are shown in \cref{fig:de}.

\begin{figure}[!htbp]
  \centering
\includegraphics[width=\linewidth]{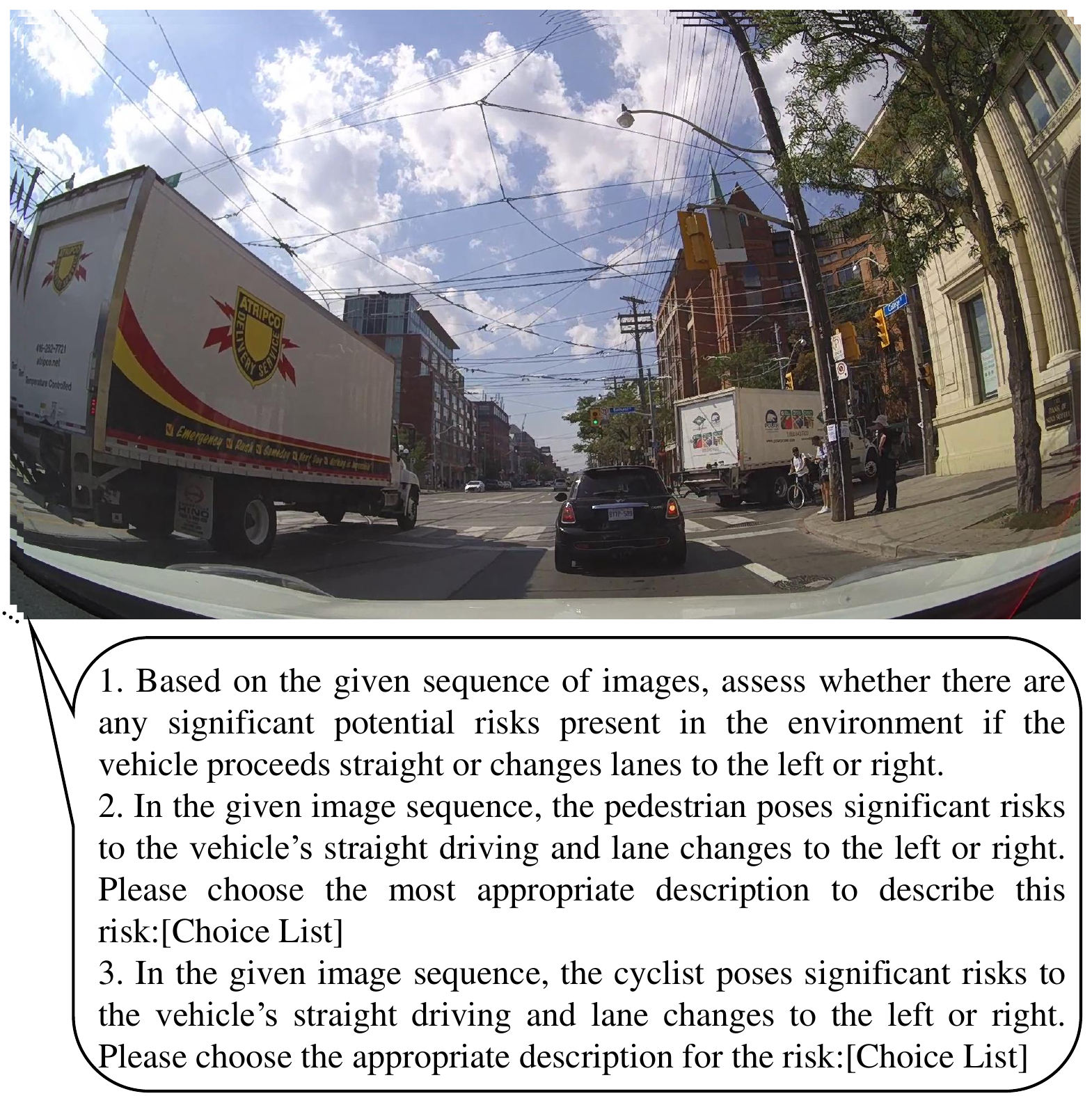}
   \vspace{-0.70cm}
   \caption{Examples of Risk Prediction.}
   \label{fig:rp}
   \vspace{-0.3cm}
\end{figure}

\noindent \textbf{Risk Prediction} evaluates the presence of significant potential risks in the environment as the vehicle proceeds straight or attempts to change lanes to the left or right. This task is divided into two steps: first, determining whether a risk exists; and second, given the source of the risk, analyzing the underlying cause of the risk. Examples are shown in \cref{fig:rp}.

\begin{figure}[!htbp]
  \centering
\includegraphics[width=\linewidth]{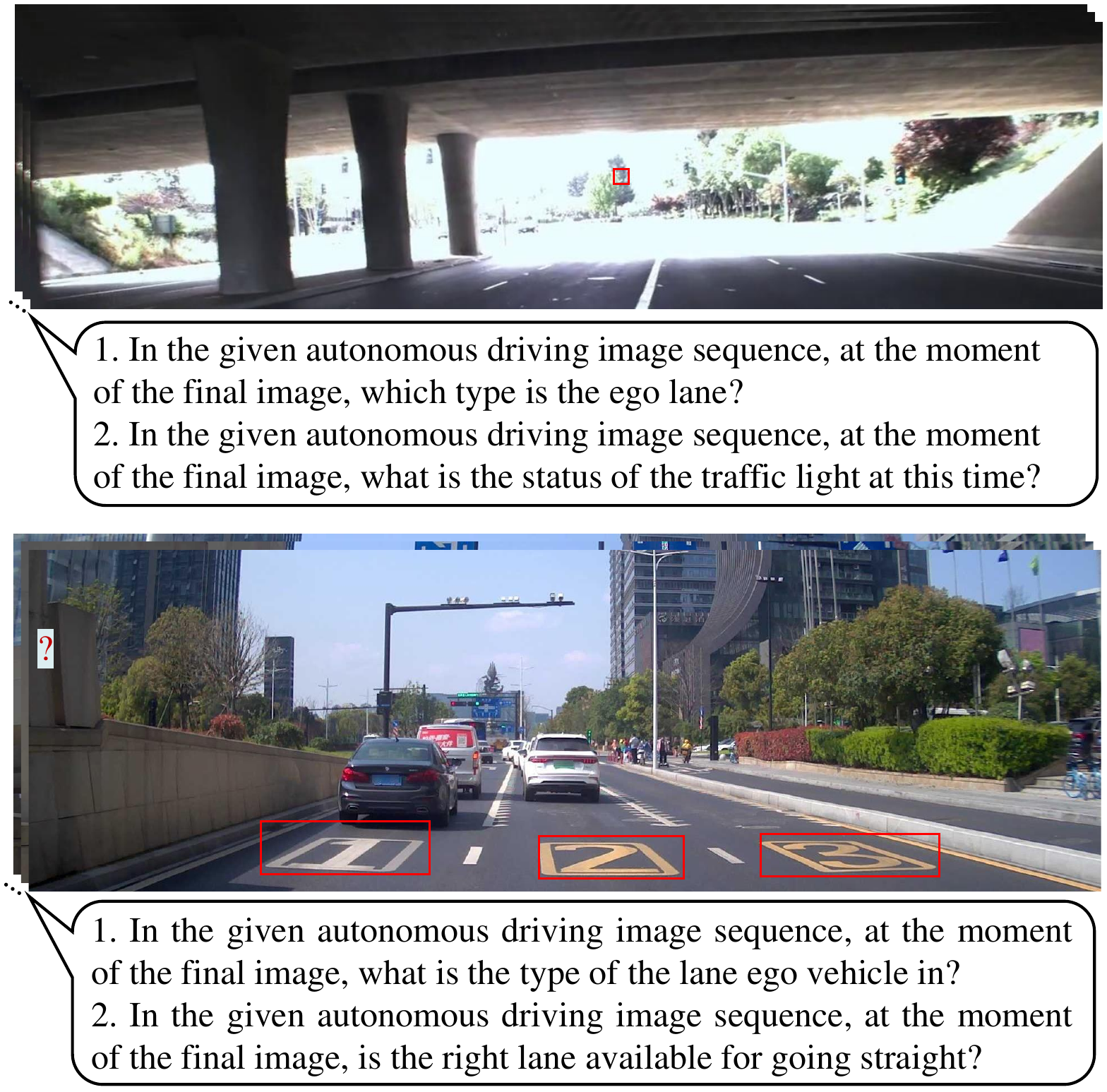}
   \vspace{-0.70cm}
   \caption{Examples of Spatio-Temporal Relation.}
   \label{fig:st}
   \vspace{-0.3cm}
\end{figure}

\noindent \textbf{Spatio-Temporal Relation} leverages information from preceding frames to infer the current driving conditions, such as unseen or occluded traffic lights and signs, or the attributes of lanes. The questions in this task are designed to require associative reasoning or recollection of previously observed information, instead of being solvable directly through simple visual cues. Examples are shown in \cref{fig:st}.

\begin{figure}[!htbp]
  \centering
\includegraphics[width=\linewidth]{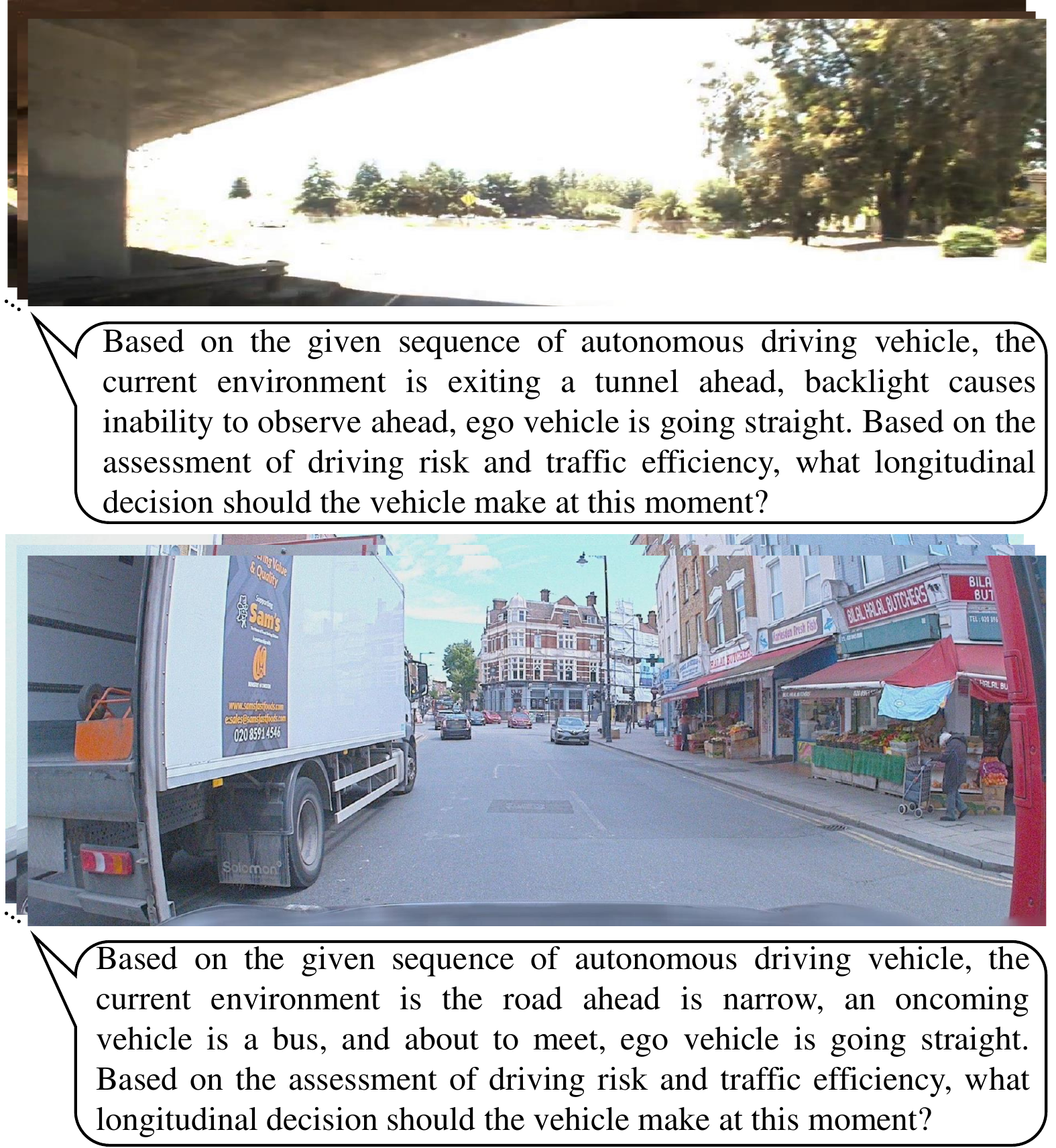}
   \vspace{-0.70cm}
   \caption{Examples of Longitudinal.}
   \label{fig:lo}
   \vspace{-0.3cm}
\end{figure}
\noindent \textbf{Longitudinal} refers to the management of a vehicle's speed and acceleration/deceleration along its direction of travel. The longitudinal operation includes maintain speed, accelerate, stop, decelerate, and decelerate to stop. Examples are shown in \cref{fig:lo}.

\begin{figure}[!t]
  \centering
\includegraphics[width=\linewidth]{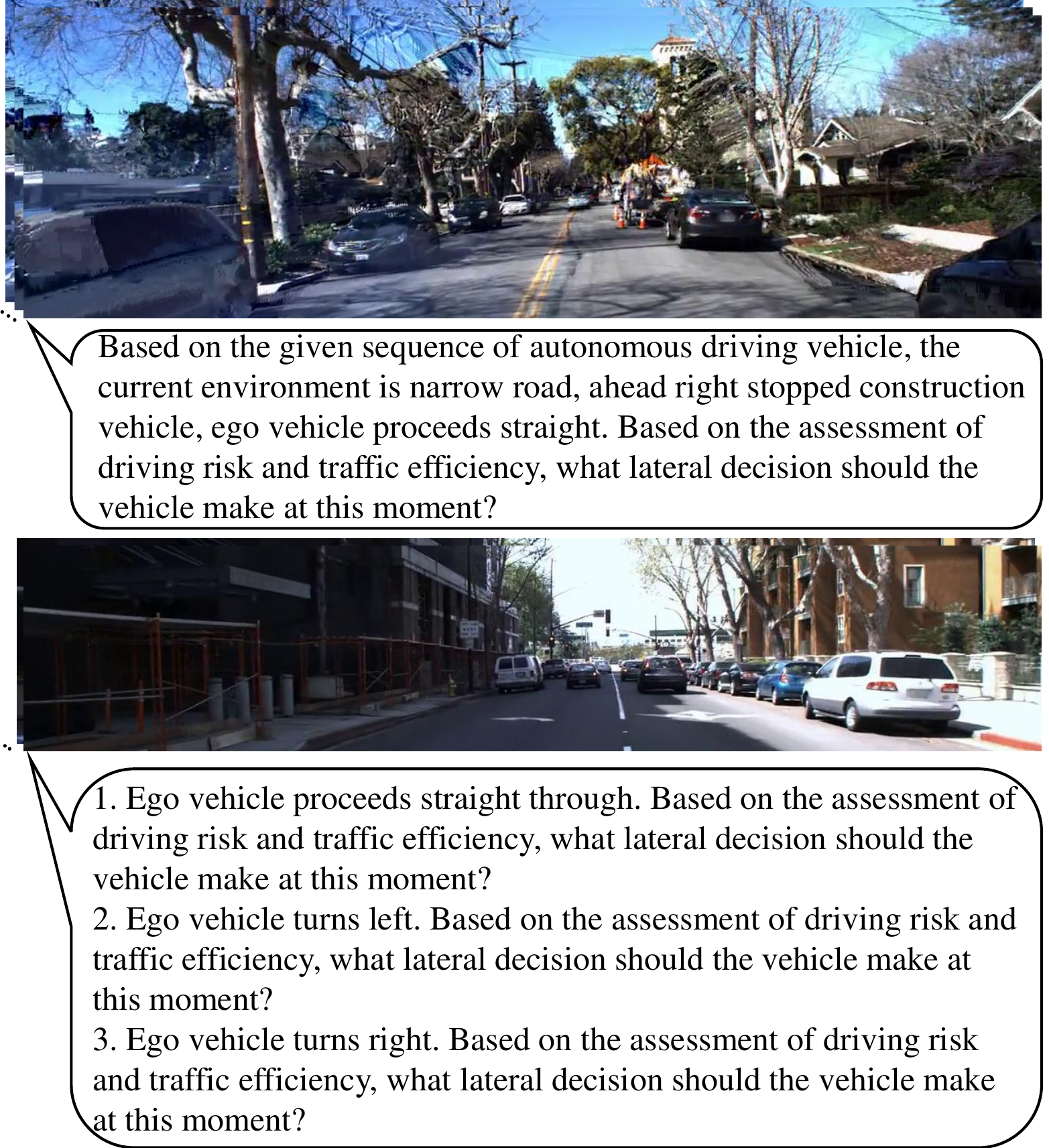}
   \vspace{-0.70cm}
   \caption{Examples of Lateral.}
   \label{fig:la}
   \vspace{-0.3cm}
\end{figure}

\noindent \textbf{Lateral} refers to the management of a vehicle's position and direction within its lane or on the road. The lateral operation includes in-lane left avoidance, in-lane right avoidance, maintain straight or change lane to the left, maintain straight or change lane to the right, borrow lane for right avoidance, borrow lane for left avoidance, change lane to the right, change lane to the left, change lane to the left or right, and maintain straight.
Examples are shown in \cref{fig:la}.

\begin{figure}[!htbp]
  \centering
\includegraphics[width=\linewidth]{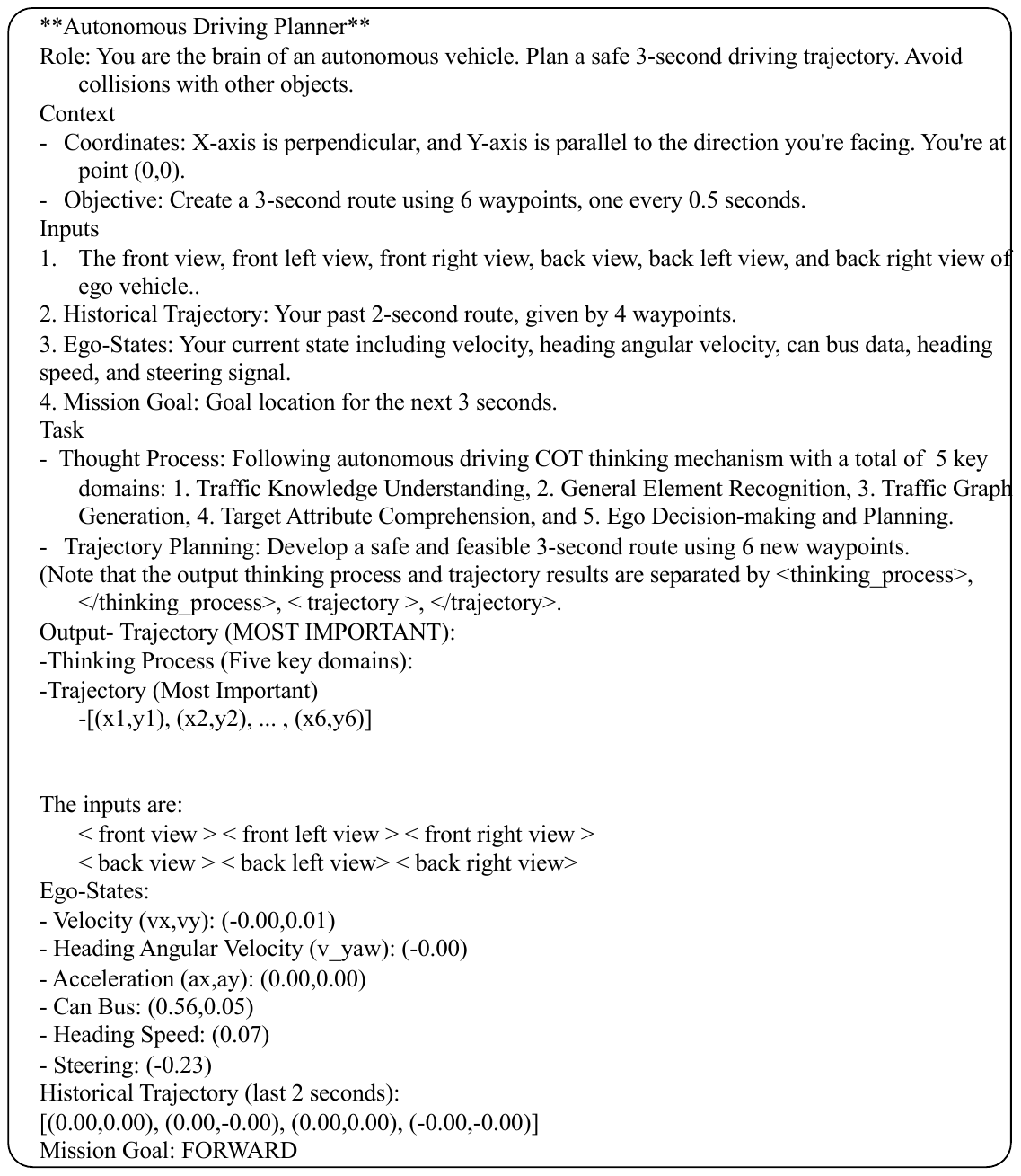}
   \vspace{-0.70cm}
   \caption{Examples of Trajectory.}
   \label{fig:tr}
   \vspace{-0.3cm}
\end{figure}

\noindent \textbf{Trajectory} prediction is formulated as a vision-language task, incorporating critical perception and prediction results along with high-level decisions.  Besides, the ego status and the historical waypoints (last 2 seconds, given by four points) are included in the instruction. The VLMs then generate a feasible 3-second driving trajectory, consisting of 6 waypoints.
An example is shown in \cref{fig:tr}.


\end{document}